\documentclass[runningheads]{llncs}

\pdfoutput=1
 
\usepackage{eccv}

\usepackage{eccvabbrv}

\usepackage{graphicx}
\usepackage{booktabs}

\usepackage[accsupp]{axessibility}  
\usepackage{multirow}
\usepackage[table,xcdraw]{xcolor}
\usepackage{float}
\usepackage{comment}
\usepackage{algorithm2e}  
\usepackage{subcaption}  

\usepackage{hyperref}

\usepackage{orcidlink}

\newif\iffinal
\iffinal

  \newcommand{\zyydelete}[1]{}

\else
    \usepackage{ulem}
    \newcommand{\stkout}[1]{\ifmmode\text{\sout{\ensuremath{#1}}}\else\sout{#1}\fi} 

    \newcommand{\zyydelete}[1]{\textcolor{red}{\stkout{#1}}}

\fi

\begin{document}

\title{SALAD: Achieve High-Sparsity Attention via Efficient Linear Attention Tuning for Video Diffusion Transformer} 





\titlerunning{SALAD}




\author{
Tongcheng Fang$^{1,2,*,\dagger}$ \quad Hanling Zhang$^{3,*}$ \quad Ruiqi Xie$^{1,*}$ \quad Zhuo Han$^1$ \quad \\ Xin Tao$^2$
\quad Tianchen Zhao$^1$ \quad Pengfei Wan$^2$ \quad Wenbo Ding$^1$ \quad \\ Wanli Ouyang$^3$ \quad Xuefei Ning$^{1\ddagger}$ \quad Yu Wang$^{1\ddagger}$
}


\authorrunning{Fang et al.}

\institute{Tsinghua University \and Kling Team, Kuaishou Technology \and The Chinese University of Hong Kong}

\maketitle
\footnotetext[1]{$*$ Co-first authors.}
\footnotetext[2]{$\dagger$ This work was conducted during the author's internship at Kling Team, Kuaishou Technology.}
\footnotetext[3]{$\ddagger$ Corresponding authors: Yu Wang (yu-wang@mail.tsinghua.edu.cn), Xuefei Ning (foxdoraame@gmail.com).}

\begin{abstract}
Diffusion Transformers have demonstrated remarkable performance in video generation. However, their long input sequences incur substantial latency due to the quadratic complexity of full attention. 
Various sparse attention mechanisms have been proposed.
Training-free approaches are limited to moderate sparsity and thus yield only modest acceleration, whereas training-based methods can reach much higher sparsity but demand substantial data and computation.
In this work, we propose \textbf{SALAD}, introducing a lightweight \textbf{linear attention branch} in parallel with the sparse attention. 
Leveraging a \textbf{Multi-level Static-Dynamic Scaling Strategy} to balance the two branches, our method attains up to \textbf{90\% sparsity} and \textbf{1.52-2.03$\times$ inference speedup} across different models and sequence lengths, while maintaining generation quality comparable to the full attention baseline.
Moreover, our finetuning process is highly efficient, requiring only \textbf{2{,}000 video samples}, \textbf{fewer than 1{,}600 training steps}, and \textbf{no more than 30 GPU hours} with a batch size of 8.
\end{abstract}

\section{Introduction}
\label{sec:intro}


Transformers face efficiency challenges in modeling long sequences due to the quadratic complexity of the attention~\cite{vaswani2017attention}. Especially for the video diffusion transformer~\cite{wan2025,yang2024cogvideox}, there is a challenge when generating high-resolution and long-duration videos. To reduce the cost of video generation, various sparse attention mechanisms have been proposed~\cite{xi2025sparse, yang2025sparse, zhang2025spargeattn, zhao2025paroattention, wu2025vmoba, zhang2025vsa, zhang2025ditfastattnv2}. Leveraging the inherent locality and sparsity of the attention map, these methods restrict each query to attend a subset of keys and values, thereby reducing the computation.

\begin{figure}[t]  
  \centering
  \includegraphics[width=0.9\columnwidth]{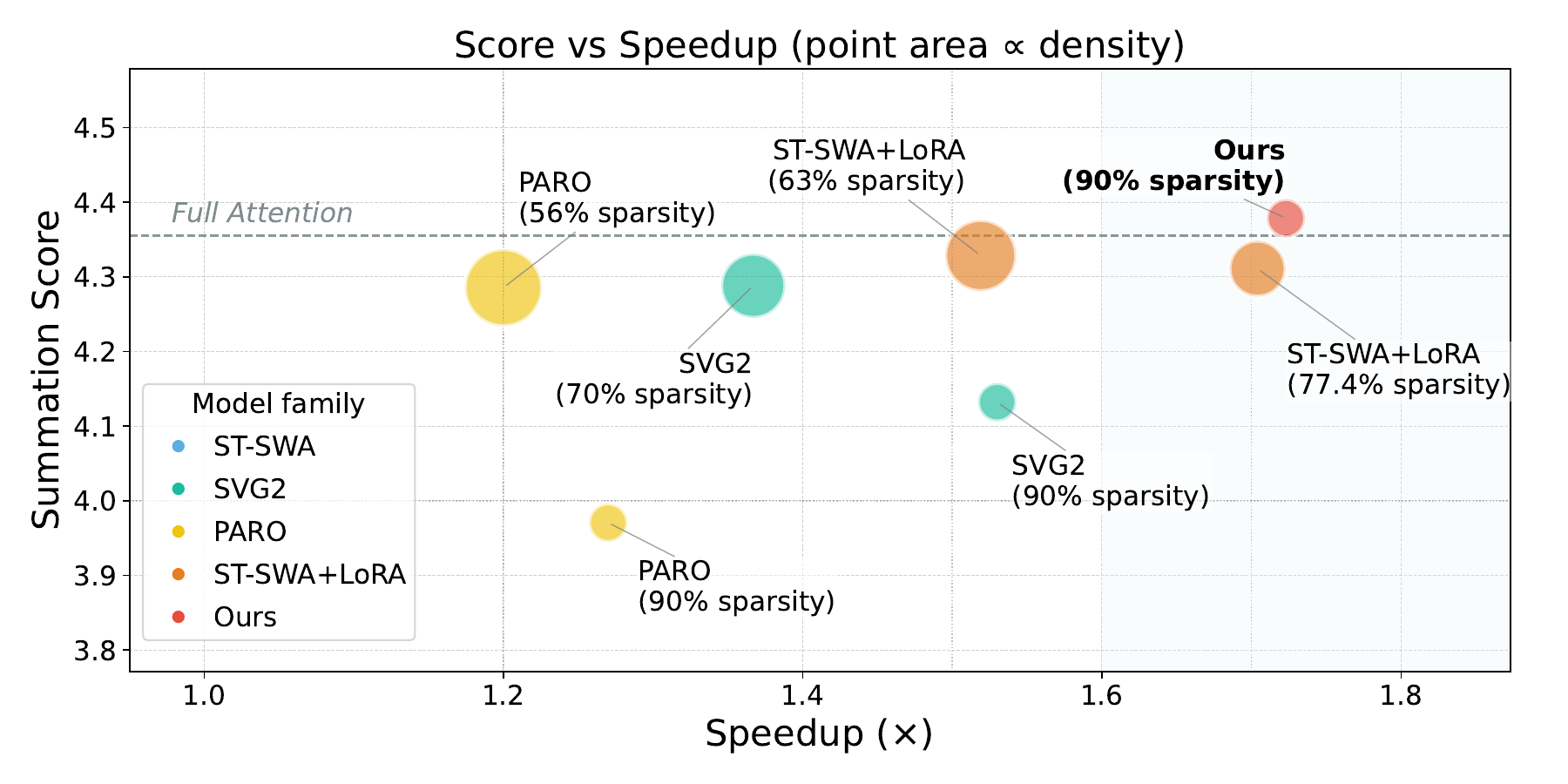}
  \caption{\textbf{Comparison of SALAD and Other Sparse Attention Mechanisms on Wan1.3B~\cite{wan2025} with 30k tokens length.} VBench Score versus speedup, with point size representing density—smaller points indicate lower computational density. This reflects overall quality–efficiency trade-off. Models compared include our approach, SVG2~\cite{yang2025sparse}, PARO~\cite{zhao2025paroattention}, and ST-SWA~\cite{zhang2025ditfastattnv2} + LoRA~\cite{hu2022lora}.}
  \label{fig:main_results}
\end{figure}


Higher sparsity leads to lower computational cost in sparse attention. However,  training-free sparse attention can only achieve limited sparsity~\cite{xi2025sparse, yang2025sparse, zhang2025spargeattn, zhang2025ditfastattnv2}, such as $40\% \sim 60\%$ under sequence length of 30k.  
Training-based approaches can achieve remarkable sparsity of $80\%$ to $95\%$, yet they incur substantial training overhead in both data preparation and computation. For instance, VMoBA~\cite{wu2025vmoba} requires approximately 182\,GPU hours to train their sparse attention model on the large-scale Koala-36M dataset~\cite{wang2025koala}. 

Fine-tuning with Low-Rank Adaptation (LoRA)~\cite{hu2022lora} provides a parameter-efficient approach for recovering performance after compression of large models~\cite{dettmers2023qlora,zhang2023loraprune}. In Transformers, LoRA typically adapts the attention projection matrices ($W^Q$, $W^K$, $W^V$, $W^O$) while keeping all other parameters frozen. But when applied to ultra-sparse attention models and fine-tuned with a limited training resources, achieving performance comparable to the dense counterpart becomes challenging. As shown in Fig.~\ref{fig:main_results}, the sparse attention model fine-tuned with LoRA still exhibits degradation relative to the full-attention baseline, even under a moderate sparsity.
The evaluation score is computed as the sum of selected VBench~\cite{huang2024vbench} metrics, including \textit{Subject Consistency}, \textit{Background Consistency}, \textit{Imaging Quality}, and \textit{Text Consistency (Overall Consistency)}.



\begin{figure}[t]
  \centering
  \includegraphics[width=0.80\columnwidth]{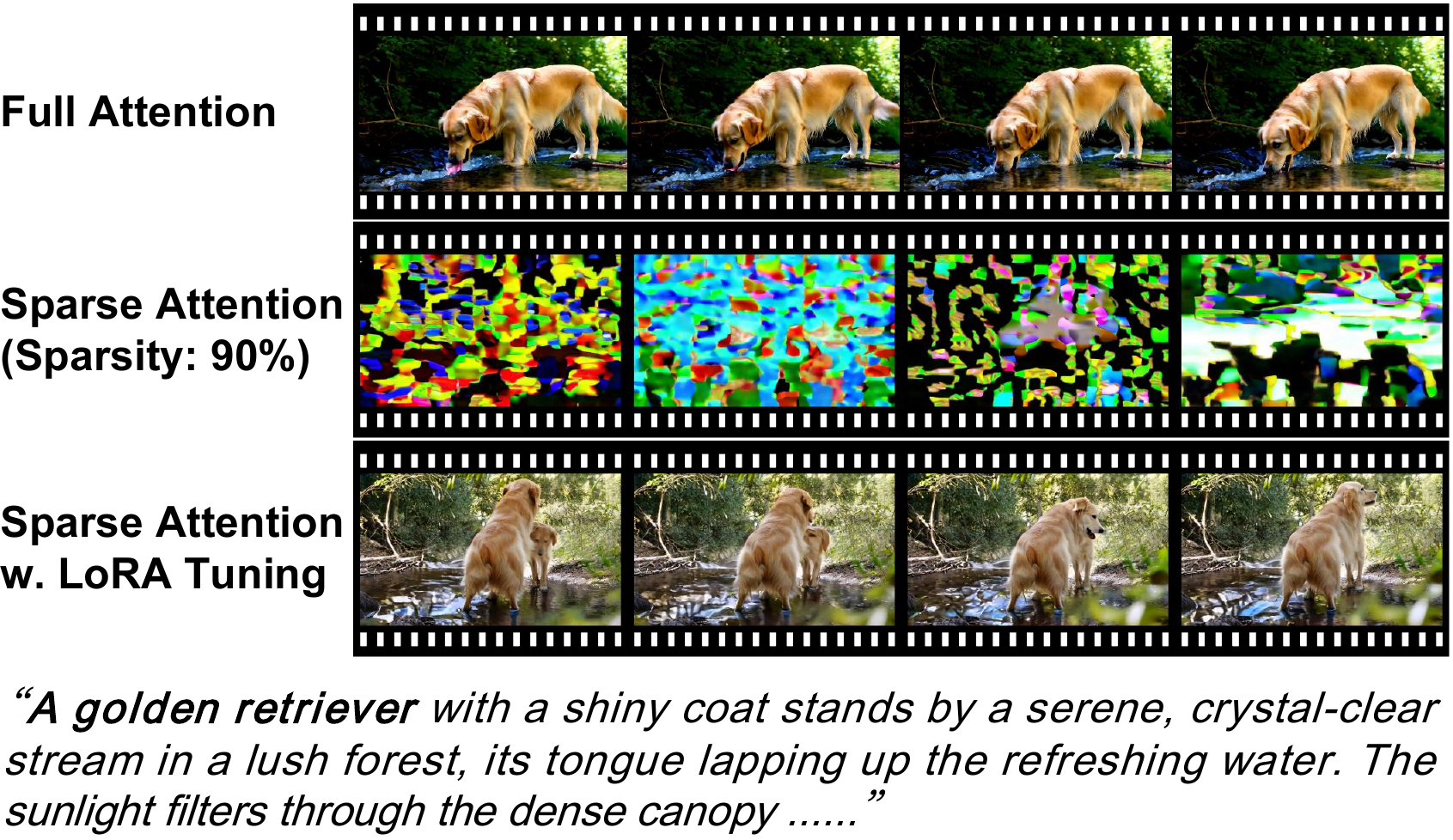}
  \caption{\textbf{Comparison of Full Attention Model, Sparse Attention Model and Sparse Attention Model with LoRA Tuning.}}
  \label{fig:failed_videos}
\end{figure}

Fig.~\ref{fig:failed_videos} presents the results generated by the sparse attention model (specifically using a sliding-window mask), along with those from the full-attention and LoRA-tuned variants. 
While LoRA fine-tuning substantially improves the generative quality of the sparse attention model, the videos still exhibit artifacts inherently tied to the sparse pattern.
For example, although the text prompt describes a dog, the LoRA-tuned model initially generates two dogs within the same frame, which gradually merge into one as the video progresses. 
Such textual and temporal inconsistencies highlight the limited effectiveness of tuning sparse attention relies solely on LoRA.
Intuitively, these artifacts arise from the intrinsic limitations of sparse attention, which restricts cross-token interactions and leads to information loss. 
For example, sliding-window attention constrains each token to attend only to its local neighbors within a single layer. 
Even with multiple stacked layers, the receptive field cannot effectively expand to cover the entire sequence~\cite{xiao2025sliding}, thereby limiting the model’s ability to capture long-range dependencies. 
This incomplete context aggregation manifests as the text and temporal inconsistencies observed in Fig.~\ref{fig:failed_videos}.

In principle, LoRA can compensate for the loss of cross-token interactions through multi-layer stacked computations. 
During fine-tuning, token representations in each attention block are refined, enabling the current layer to incorporate information from distant tokens. 
Through layer-by-layer propagation, deeper layers can thus access broader contextual information, theoretically restoring long-range interactions. 
Consequently, LoRA fine-tuning holds the potential to recover performance comparable to full attention. 
However, our experiments show that achieving substantial recovery remains difficult under extreme high sparsity.




To compensate for the information loss in sparse attention, we introduce a linear attention branch. 
This branch enables global token mixing, offering the potential to restore critical cross-token interactions that are neglected under ultra-high sparsity attention.
Moreover, its $O(N)$ computational complexity incurs minimal overhead during inference, allowing the model to retain most of the efficiency gains provided by high-sparsity attention.



However, we empirically find that the linear attention branch can only supplement limited critical information while struggling to model the entire sequence. Thus, the sparse attention should dominate the output while the linear attention branch should be set to an auxiliary role.

To enable practical sparse-linear attention, we propose \textbf{SALAD} (High-\textbf{S}parsity \textbf{A}ttention paralleling with \textbf{L}inear \textbf{A}ttention for \textbf{D}iffusion Transformer), an efficient attention module that facilitates high-sparsity modeling through an auxiliary linear attention branch. Specifically, we introduce a linear attention branch in parallel with sparse attention. We further design a \textbf{Multi-level Static-Dynamic Scaling Strategy} to regulate the linear branch across different blocks and timesteps in the Diffusion Transformer.
Our contribution can be summarized as follows:
\begin{enumerate}
\item We propose \textbf{SALAD}, an efficient attention architecture designed to enhance the performance of ultra-sparse attention through a parallel linear attention branch. 
Most parameters of the linear branch are shared with the sparse attention module, introducing only about \textbf{4.99\%} additional parameters relative to the pretrained model.

\item We observe that in hybrid sparse-linear attention, it is essential to regulate the influence of the linear attention branch. 
We design a Multi-level Static-Dynamic Scaling Strategy that precisely controls its contribution, effectively improving overall model performance.

\item 
With only \textbf{2k} open-source video samples and \textbf{less than 1.6k} training steps, \textbf{SALAD} achieves up to \textbf{90\% sparsity} and \textbf{1.52-2.03$\times$ inference speedup} across different models and sequence lengths, while maintaining generation quality comparable to that of the full attention counterpart.

\end{enumerate}

\section{Related Work}
\label{sec:related_work}
\noindent{\textbf{Video Diffusion Models.}} Diffusion Transformer~\cite{Peebles2022DiT} has demonstrated its advantage in video generation. 
Earlier approaches such as Latte~\cite{ma2024latte} decouple video sequences into spatial and temporal tokens and apply attention separately, which limits the model’s expressive capacity. 
More recent works~\cite{brooks2024video, yang2024cogvideox, wan2025} instead process the entire video as a unified token sequence and perform full attention across all tokens.
While improving generation quality, it introduces substantial memory and computational overhead due to the extended sequence length.

\noindent{\textbf{Efficient Video Diffusion Transformer.}}
To address the computational challenges of video diffusion transformers, various efficient video generation techniques have been proposed. 
Quantization~\cite{zhao2024viditq, li2024svdquant, chen2025q} reduces memory usage and inference latency by lowering the precision of model weights and activations. 
Caching mechanisms~\cite{lv2024fastercache, kahatapitiya2025adaptive, cui2025bwcache, liu2025timestep} reuse intermediate features across denoising timesteps to avoid redundant computations. 
Timestep distillation~\cite{zheng2025large} accelerates generation by reducing the number of required sampling steps. 
Token merging compresses the sequence length by merging similar tokens, thereby mitigating the memory and computational overhead of long videos~\cite{yuan2025dlfr}.
In addition, efficient attention mechanisms have been widely explored to address the long video token sequence. 
Sparse attention reduces computational cost by restricting cross-token interactions~\cite{xi2025sparse, yuan2024ditfastattn, zhang2025fast, yang2025sparse, zhao2025paroattention}. 
Linear attention has also been adopted, transforming the quadratic complexity of attention into linear complexity and thus reducing the cost of modeling long video sequences~\cite{wang2025lingen, chen2025sana}.

\begin{figure*}[t]
  \centering
  \includegraphics[width=1\textwidth]{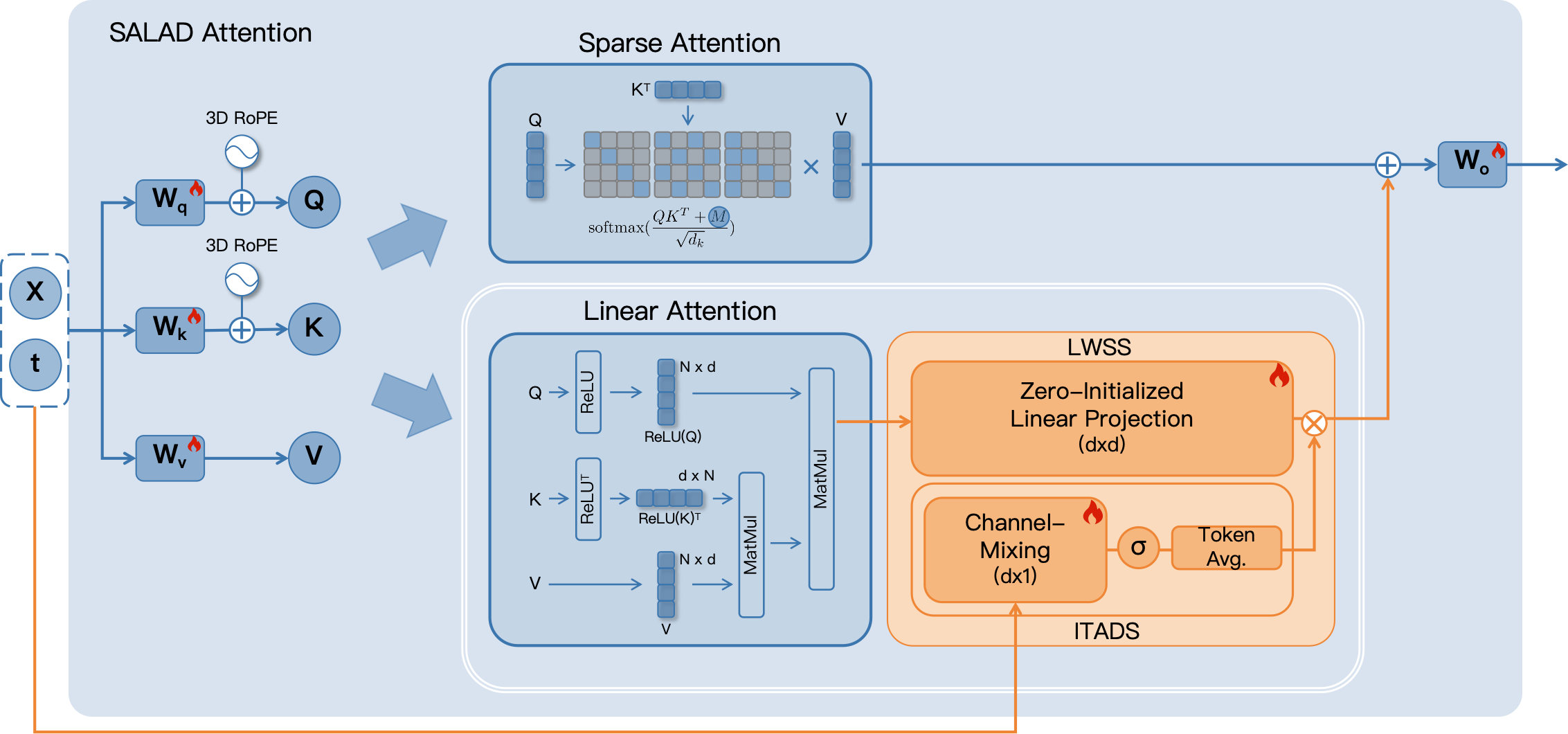}
  \caption{\textbf{Overview of SALAD Attention Module.}}
  \label{fig:salad}
\end{figure*}

\noindent{\textbf{LoRA for Model Compression Recovery.}}
LoRA~\cite{hu2022lora} is an efficient fine-tuning technique that freezes the pre-trained model weights and injects trainable low-rank decomposition matrices. 
Beyond downstream task adaptation, LoRA has also been widely adopted for model compression recovery. 
For instance, Ma et al.~\cite{ma2023llm} combine LoRA with Large Language Models pruning to mitigate performance degradation with minimal fine-tuning overhead. 
QLoRA~\cite{dettmers2023qlora} introduces quantized LoRA tuning for 4-bit model performance recovery, while LongLoRA~\cite{chen2023longlora} extends LoRA to normalization, embedding, and Feed-Forward Network layers for long-context adaptation under local attention.
In this work, we find that LoRA fails to fully recover the performance of ultra-sparse Video Diffusion Transformers and propose SALAD as a more effective tuning approach.



\section{Preliminary}
\label{sec:preliminary}
\noindent \textbf{Sparse Attention.}
Given queries ($Q$), keys ($K$), and values ($V$) $\in \mathbb{R}^{N \times d}$, the full attention mechanism can be formalized as follows:

\begin{equation}
\operatorname{Attention}(Q, K, V) = \operatorname{softmax}\left(\frac{QK^T}{\sqrt{d_k}}\right) V.
\label{eq:attention}
\end{equation}
The computation of the attention matrix $QK^\top$ incurs $O(N^2 d)$ memory and computation complexity, which is prohibitive for large $N$.

To reduce the high computational cost, sparse attention limits the interactions to only a selected group of token pairs. This is formally done by incorporating a sparsity mask \( M \in \{-\infty, 0\}^{n \times n} \) into the attention operation: 
\begin{equation}
\mathrm{SparseAttention}(Q, K, V) = \operatorname{softmax} \left( \frac{QK^\top + M}{\sqrt{d_k}} \right) V.
\label{eq:sparse_attention}
\end{equation}
Positions marked with $-\infty$ are effectively excluded from the softmax operation.

Sparse attention mechanisms can be categorized into \textit{static} and \textit{dynamic} approaches, depending on how the sparse mask is obtained.

Static sparse attention predefined the sparse mask prior to inference, often through heuristic or data-driven calibration strategies~\cite{yuan2024ditfastattn, zhang2025ditfastattnv2, zhao2025paroattention} that exploit prior knowledge about attention patterns.
In contrast, dynamic approaches determine the sparse mask on-the-fly during inference, allowing the sparsity pattern to adapt to the input data via learned or rule-based decision functions~\cite{xi2025sparse, yang2025sparse}.




Besides, the sparse mask can take different structural forms. 
In this work, we explore both \textit{static} and \textit{dynamic} sparse attention mechanisms, specifically the static \textit{sliding-window attention (SWA)} and the dynamic \textit{Top-$k$ attention}.  

\textit{Sliding-window attention (SWA)} exploits the inductive bias of locality within the input sequence~\cite{yuan2024ditfastattn, zhang2025fast}. 
Its sparse mask can be efficiently derived through calibration on a small subset of data. 
Motivated by ~\cite{xi2025sparse}, we further apply spatial-temporal reordering on this design to better handle video tokens, termed \textbf{ST-SWA}. 
Details of it are provided in the Appendix~\ref{app:sparse_details}.

\textit{Top-$K$ attention} dynamically constructs irregular sparse masks by retaining the most informative query–key pairs.  
For each query, only the $k$ keys with the highest importance scores are preserved, allowing the model to focus computation on salient interactions.  
In this work, we use Top-K sparse attention implementation of VMoBA~\cite{wu2025vmoba}. In the meantime, \textit{Top-$P$ attention} adopts a similar principle but retains a variable number of keys whose cumulative importance exceeds a threshold $p$~\cite{lu2025moba, yuan2025native, zhang2025sla}.

\noindent \textbf{Linear Attention.}
Linear attention reduces the computational complexity from $O(N^2 d)$ in standard attention to $O(N d^2)$ by replacing the softmax operation with a kernel function $\phi(\cdot)$ applied to $Q$ and $K$. This allows the attention weights to be approximated as $\phi(Q)\phi(K)^\top$, enabling associative reordering. Specifically, one can first compute $H = \phi(K)^\top V$ and $Z = \phi(K)^\top \mathbf{1}$, and then obtain the output as $O = \frac{\phi(Q) H}{\phi(Q) Z}$.
A common choice for $\phi(\cdot)$ is ReLU function~\cite{xie2024sana, chen2025sana}. The relu-based linear attention can be formulated as:

\begin{align}
O_i &= 
\frac{\mathrm{ReLU}(Q_i)
\left( \sum_{j=1}^{N} \mathrm{ReLU}(K_j)^\top V_j \right)}{
\mathrm{ReLU}(Q_i)
\left( \sum_{j=1}^{N} \mathrm{ReLU}(K_j)^\top \right)}.
\label{eq:relu_linear_attention}
\end{align}


\section{Methods}
\label{sec:methods}





The architecture of SALAD is illustrated in Fig.~\ref{fig:salad}. 
To achieve high-sparsity attention in video diffusion transformers while preserving generation quality, SALAD augments sparse attention with a lightweight parallel linear attention branch. Both branches share the same query ($\mathbf{Q}  $), key ($\mathbf{K}$), and value ($\mathbf{V}$). The projection matrices $W^Q$, $W^K$, $W^V$, and $W^O$ are initialized from the pretrained weights and fine-tuned using LoRA during training. The linear attention output is residually added to the sparse output after multi-leval scaling.

In this work, we observe that carefully regulating the output range of linear attention is critical to overall performance.
Thus, we introduce a multi-level static-dynamic scaling strategy that finely balances the linear branch's contribution. This comprises (1) a layer-wise static scaling implemented via a zero-initialized projection layer
and (2) a lightweight input- and timestep-aware dynamic scaling module that modulates the linear influence based on the denoising timestep $t$ and current hidden states. 
Together, these mechanisms ensure the linear branch acts as a controlled complement, efficiently injecting missed information under extreme sparsity without overwhelming sparse attention output. The complete SALAD attention computation is presented in Alg.~\ref{alg:salad_attention}.

SALAD can be generally applied to various sparse attention mechanisms, including both static and dynamic variants. In our experiments, we adopt a modified sliding-window attention~\cite{yuan2024ditfastattn} and a self-implemented top-$k$ sparse attention~\cite{wu2025vmoba}, as described in Sec.~\ref{sec:preliminary}. These two approaches are the static and dynamic sparse attention, respectively. For the linear attention branch, we use a ReLU-based linear attention. The linear attention branch incorporates 3D Rotary Position Embeddings~\cite{su2024roformer} to model spatial-temporal dependencies.

\RestyleAlgo{ruled} 

\begin{algorithm}[t]
    \caption{SALAD Attention with Multi-level Scaling}
    \label{alg:salad_attention}
    \KwIn{$X$: Unnormalized hidden states; $t$: Denoising timestep}
    \KwOut{$O$: Output of the SALAD attention block}

    $e_t = \operatorname{Time Emb}(t)$ 
    \tcp{Time embedding inherited from the pretrained DiT}
    
    $X_t = \operatorname{Modulation}(\operatorname{norm}
    (X), e_t) $
    
    \tcp{Feature Transformation}
    $\mathbf{Q} = X_t W^Q, \quad \mathbf{K} = X_t W^K, \quad \mathbf{V} = X_t W^V$\;
    
    $O_{s} = \operatorname{SparseAttn}(\mathbf{Q}, \mathbf{K}, \mathbf{V})$ 
    
    $O_{l} = \operatorname{LinearAttn}(\mathbf{Q}, \mathbf{K}, \mathbf{V})$
    
    \BlankLine
    
    \tcp{Level 1: Static Layer-wise Scaling}
    $O'_{l} = \operatorname{Proj}(O_{l})$ \tcp*{Zero-initialized linear layer}
    
    \tcp{Level 2: Dynamic Input-Timestep-Aware Scaling}
    
    $s = \operatorname{Mean}(\sigma(\operatorname{Linear}(X_t))$ \tcp*{Global scaling factor}
    
    \BlankLine
    \tcp{Residual Integration and Final Projection}
    $\tilde{O} = O_{s} + s \cdot O'_{l}$ 
    
    $O = \tilde{O} W^O$\;
    
\end{algorithm}



\subsection{Rank Imbalance between Linear and Sparse Branches}
\label{sec:linear_for_sparse}

\begin{figure}[t!]
    \centering
    
    \begin{minipage}{0.45\textwidth}
        \centering
        \includegraphics[
            width=\linewidth,
            keepaspectratio,       
            ]{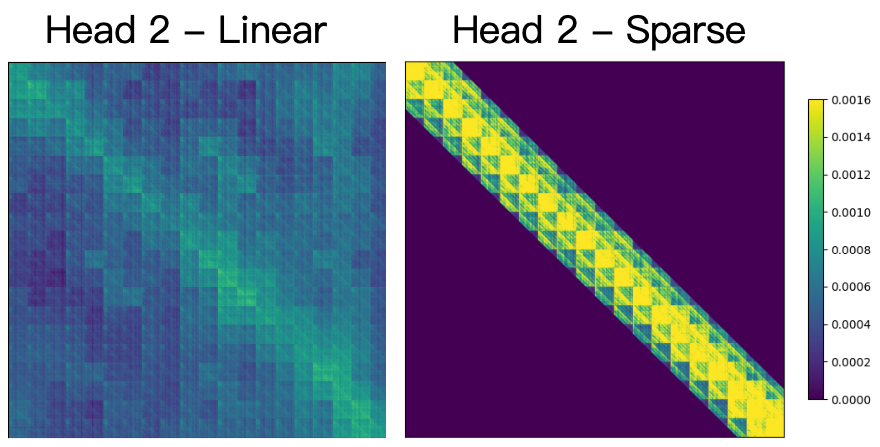}
        \caption{Linear and sparse attention maps at block 8, timestep 20, head 2. The linear attention can achieve global token interaction, whereas sparse attention restricts each query to interact with only a subset of tokens.}
        \label{fig:attn_map}
    \end{minipage}\hfill
    \begin{minipage}{0.49\textwidth}
        \centering
        \includegraphics[
            width=\linewidth,
            height=5.8cm,          
            keepaspectratio,
            ]{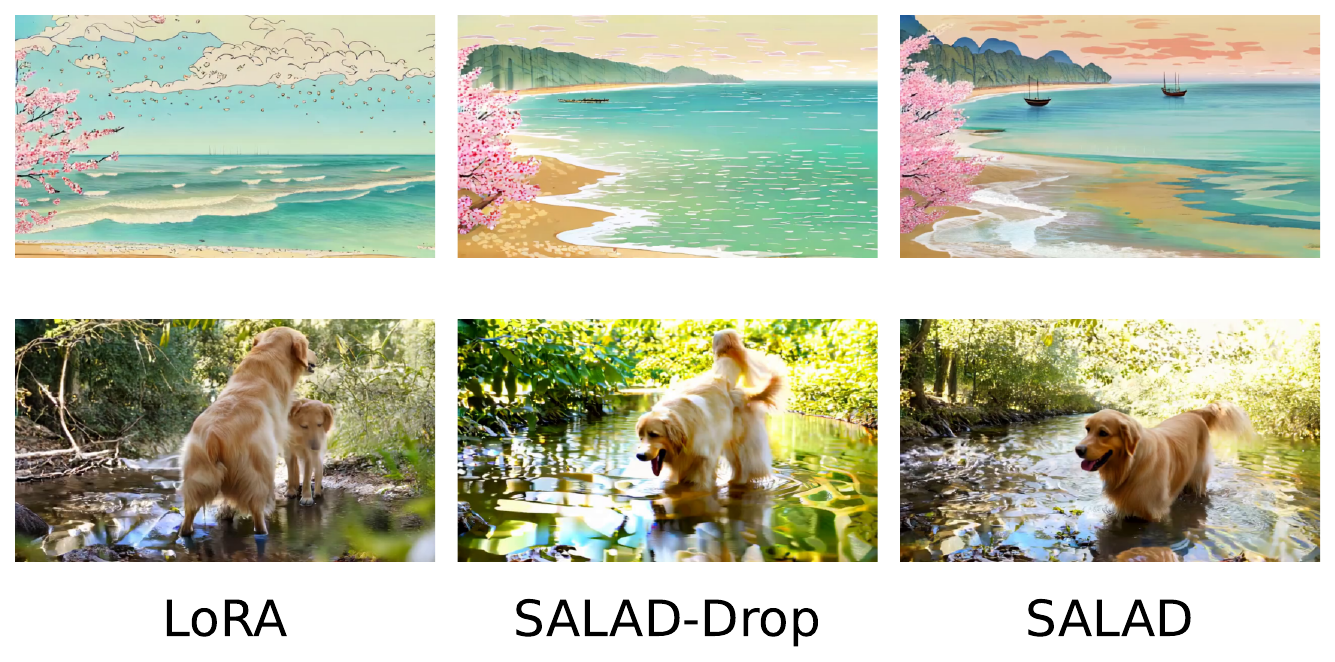}
        \caption{Generated Video Samples of sparse attention tuned with LoRA, SALAD, and SALAD with linear branch dropped during inference.}
        \label{fig:branch_analysis}
    \end{minipage}
    
\end{figure}
\begin{figure}[t!]
    \centering
    \begin{subfigure}{0.495\columnwidth}
        \centering
        \includegraphics[width=\linewidth]{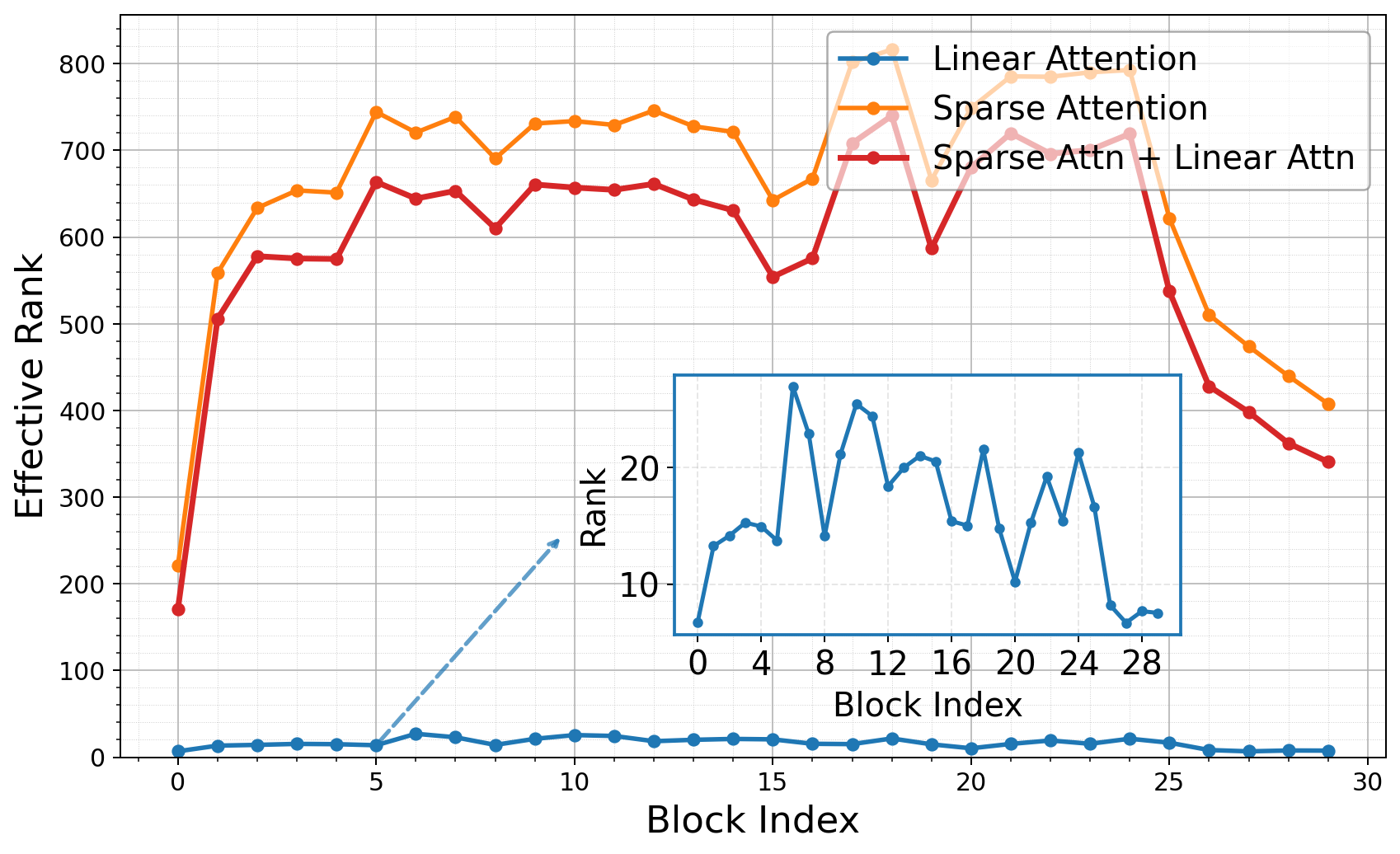}
    \end{subfigure}
    \hfill
    \begin{subfigure}{0.49\columnwidth}
        \centering
        \includegraphics[width=\linewidth]{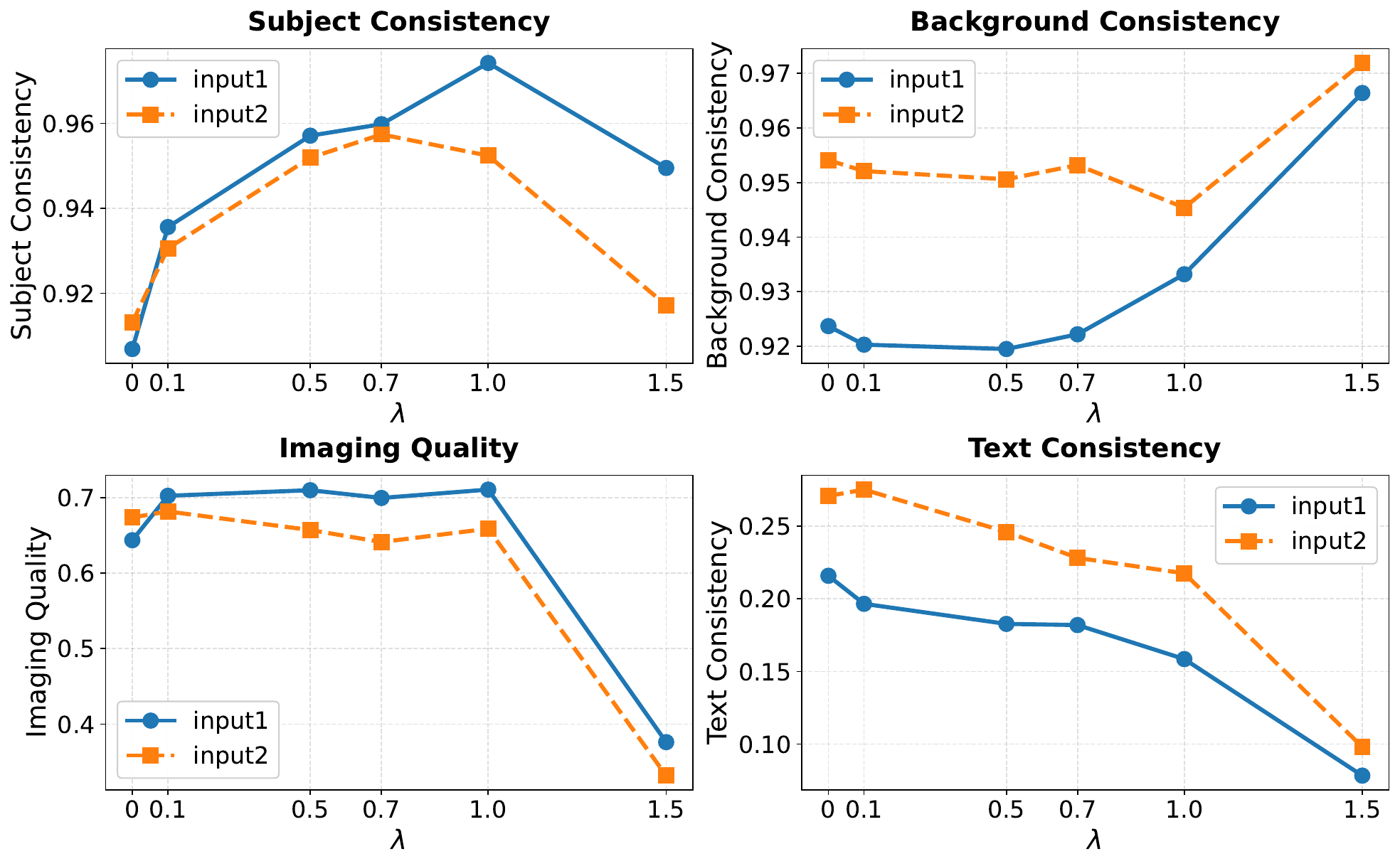}
    \end{subfigure}
    
    \caption{\textbf{Effective rank comparison and effect of linear attention scaling.} 
    (Left) Effective rank of sparse attention and linear attention outputs. 
    (Right) As $\lambda$ decreases, we observe improvements in some dimensions. Moreover, the optimal value of $\lambda$ varies across different inputs. These observations suggest that relying solely on a static projection layer to regulate the influence of the linear attention branch is insufficient. }
    \label{fig:rank_and_branch_effect}
\end{figure}

A straightforward way to enhance sparse attention with linear attention is to directly combine their outputs by summation and fine-tune the model. However, we empirically observe that this integration leads to suboptimal performance. 

we analyze the attention maps and effective rank of the two branches. As shown in Fig.~\ref{fig:attn_map}, the linear attention branch indeed captures global token interactions, suggesting its potential to compensate for information missed by highly sparse attention. Nevertheless, examining the effective rank of the branch outputs uncovers a significant imbalance. As illustrated in Fig.~\ref{fig:rank_and_branch_effect}~(Left), the linear attention output exhibits a substantially lower rank than that of sparse attention, indicating a pronounced capacity gap between the two branches.

More importantly, we find that directly adding the two branches can even reduce the overall output rank compared to its of sparse attention. This suggests that the low-rank linear branch may inadvertently suppress informative high-rank structures from the sparse attention when its magnitude is not properly controlled. These observations indicate that the linear attn branch should serve as a auxiliary complement rather than a co-equal contributor. If over-amplified, it may overwhelm sparse attention output and degrade the modeled information.

\subsection{Multi-level Static-Dynamic Scaling for Linear Attention Branch}
\label{sec:multi_level_scaling}
The rank imbalance analysis indicates that the sparse branch carries the primary sequence modeling load, while the linear branch functions as an auxiliary path, mainly compensating for information missed by highly sparse attention.
We speculate that the low-rank linear attention component may suppress or overwhelm informative high-rank structures from the sparse branch without explicit regulation. Therefore, properly down-weighting the linear branch, rather than allowing it to dominate or vanish is crucial.
To this end, we introduce a \textbf{multi-level static-dynamic scaling strategy}, consisting of a layer-wise static scaler and an input-timestep-aware dynamic scaler.

\noindent\textbf{Layer-wise Static Scaling (LWSS).}
As shown in Fig.~\ref{fig:rank_and_branch_effect}, the degree of rank degradation varies across different blocks. And different transformer layers exhibit varying sensitivity to sparse attention~\cite{yuan2024ditfastattn, zhang2025ditfastattnv2}, and consequently require different degrees of compensation from the linear attention branch. 
To accommodate this heterogeneity, we introduce a layer-wise static scaling mechanism that adaptively down-weights the contribution of linear attention at each layer. 

Experiments reveal that inserting a learnable projection layer after the linear attention branch in each Transformer layer helps preliminarily regulate the contribution of linear attention.
And inspired by the LoRA~\cite{hu2022lora}, the projection layer is zero-initialized. It allows the model to start tuning from the post-training sparse attention, ensuring more stable tuning in the early stage. Meanwhile, this design encourages the linear attention branch to be progressively incorporated from a small-magnitude regime, enabling and gradual increase of its contribution.
The projection layer is trained during fine-tuning and is formulated as:
\begin{equation}
    O = O_s + \operatorname{ZeroInitProj}(O_l),
\end{equation}
where $O_s$ and $O_l$ denote the outputs of the sparse and linear attn, respectively. 

However, when introducing a scaling hyperparameter $\lambda$,
\begin{equation}
O = O_s + \lambda \cdot \operatorname{ZeroInitProj}(O_l),
\label{eq:lambda_scale}
\end{equation}
we find that the performance can be further improved.  As shown in Fig.~\ref{fig:rank_and_branch_effect}~(Right), although models are trained with $\lambda=1$, moderately reducing $\lambda$ during inference often yields better results, while excessively small values degrade performance. 
This phenomenon suggests that the linear branch indeed requires more careful magnitude control. Relying solely on a static scaling strategy appears insufficient, highlighting the need for more dynamic control.
However, both the projection layer and the hyperparameter $\lambda$ are static. They cannot adapt to different inputs or diffusion timesteps, resulting in a suboptimal performance.

    

\noindent\textbf{Input-Timestep-Aware Dynamic Scaling (ITADS).}
To enable adaptive regulation, we introduce an input-timestep-aware dynamic scaling module.
As depicted in Alg.~\ref{alg:salad_attention} and Fig.~\ref{fig:salad}, the input fused with the timestep embedding is passed to a learnable linear layer with channel mixing, followed by a nonlinear function. Then token averaging is performed to produce a global scaling factor.

Since scaling factors $s$ smaller than one empirically lead to better behavior, we adopt a sigmoid activation to constrain $s \in (0,1)$. The dynamic scaling module introduces minimal overhead and is fully trainable during fine-tuning.

Together, the static and dynamic scaling mechanisms ensure that the linear branch remains an auxiliary low-rank complement, while adaptively injecting global information when necessary.
After applying the proposed scaling strategy, we observe that the rank of the combined output remains comparable to sparse attention output.
Moreover, we observe a lower subspace overlap between the linear and sparse branches, measured by the Frobenius norm of the projection between their top principal components obtained via PCA~\cite{mackiewicz1993principal}. 
This suggests that the two branches learn more complementary representations, enabling the linear branch to effectively compensate for information missing in sparse attention. More details are provided in the Appendix~\ref{app:rank_analysis}.

\subsection{Post-tuning Branch Dropping}
\label{sec:post_branch_drop}
The contribution of the linear attention branch varies across timesteps and inputs, as reflected by the dynamic scaling factors. 
We observe that branches with small scaling values contribute marginally and can be dropped without degrading performance.
Based on this observation, we introduce a simple branch-dropping strategy guided by the scaling factors. 

Specifically, we calibrate on a small validation prompt set to determine a timestep-wise threshold $\tau_t$. 
During inference, branches with global scaling factor below $\tau_t$ are dynamically pruned. More details are provided in the Appendix~\ref{app:branch_drop}.

Interestingly, dropping a small portion of branches with extremely low scaling values can even slightly improve generation quality (see Table~\ref{tab:salad_result_wan14b}). 
We conjecture that some layers at a certain timestep are naturally better suited for sparse attention, such that the linear branch may introduce minor negative effects.


\section{Experiments}
\label{sec:experiments}



\begin{table*}[t]
\centering
\caption{\textbf{Quality and efficiency results of SALAD and baselines on WAN1.3B.} * indicates the sparse attention is applied to a part of diffusion steps. Following the SVG2~\cite{yang2025sparse} practice “with sparse attention skipped during the first 30\% of denoising steps for all methods, as these steps are critical for generation quality”, several baselines adopt the same setting.}
\label{tab:main_result}
\resizebox{0.98\textwidth}{!}{
\begin{tabular}{@{} c c c ccccc cc @{}}
\toprule
\textbf{Tokens} & \textbf{Method} & Category & SC↑ & BC↑ & IQ↑ & TC↑ & VR↑ & Sparsity↑ & Speedup↑ \\
\midrule
30k & Original (Full Attention) & — & 95.88 & 96.17 & 65.93 & 25.31 & 0.109 & 0\%  & — \\
\midrule
30k & Spatial-Temporal          & Training-free  & 80.14 & 93.96 & 42.07 & 7.98  & -0.200 & 90\%* & 1.52$\times$ \\
30k & SVG2                      & Training-free  & 87.29 & 93.76 & 60.66 & 25.01 & -0.036  & 90\% & 1.53$\times$ \\
30k & PAROAtten                 & Training-free  & 83.07 & 88.16 & 57.51 & 24.96 & -0.072  & 90\% & 1.27$\times$ \\
\midrule
30k & SLA                       & Tuning-based   & 83.77 & 96.26 & 25.82 & 3.60  & -0.205 & 90\% & 1.42$\times$ \\
30k & Top-K w.\ LoRA            & Tuning-based   & \underline{96.84} & 96.10 & \underline{66.03} & 24.02 & 0.058  & 80\% & 0.95$\times$ \\
\rowcolor{gray!12}
30k & Top-K w.\ SALAD           & Tuning-based   & \textbf{97.15} & \underline{96.27} & 64.76 & \textbf{25.83} & \textbf{0.107} & 90\% & 1.04$\times$ \\
30k & Spatial-Temp.\ w.\ LoRA   & Tuning-based   & 94.96 & 95.29 & 65.27 & 25.39 & 0.072 & 90\%* & 1.52$\times$ \\
\rowcolor{gray!12}
30k & Spatial-Temp.\ w.\ SALAD  & Tuning-based   & 96.54 & \textbf{96.37} & \textbf{66.09} & \underline{25.55} & \underline{0.092} & 90\% & 1.72$\times$ \\
\midrule\midrule
100k & Original (Full Attention) & —              & 92.09 & 95.47 & 65.29 & 27.50 & 0.112      & 0\%  & — \\
\midrule
100k & SVG2                      & Training-free  & 94.92 & 95.45 & 65.02 & 26.07 & 0.062      & 70\%* & 1.71$\times$ \\
100k & SVG2                      & Training-free  & 92.65 & 95.18 & 62.02 & 25.38 & 0.027      & 90\% & 1.96$\times$ \\
\midrule
100k & Spatial-Temp. w.\ LoRA           & Tuning-based   & 95.27 & 95.78 & 61.93 & 25.79 & 0.064      & 90\%* & 1.99$\times$ \\
\rowcolor{gray!12}
100k & Spatial-Temp. w.\ SALAD          & Tuning-based   & \textbf{96.43} & \underline{95.58} & \underline{64.46} & \textbf{26.10} & \textbf{0.085}      & 90\% & 2.03$\times$ \\
\bottomrule
\end{tabular}
}
\end{table*}

\subsection{Setup}
\noindent\textbf{Base Model \& Datasets.} 

We evaluate our method against representative baselines on Wan2.1-1.3B~\cite{wan2025} at 480p with 77 frames (30k tokens).
We also extend Wan-1.3B to 720p (100k tokens) with SALAD, and apply our method to Wan2.1-14B at 480p. 
All training-based methods are fine-tuned on 2,000 Mixkit videos under the OpenSora plan~\cite{lin2024open}.

\noindent\textbf{Baselines.} 
We compare our method with various efficient video generation approaches. 
For training-free baselines, we include two static sparse attention methods: spatio-temporal sliding window attention, PAROAttention~\cite{zhao2025paroattention}
and SVG2~\cite{yang2025sparse}. 
The spatio-temporal local attention is extended from DiTFastAttnV2, incorporating the token reordering kernel from SVG. 
Following common practice~\cite{xi2025sparse, yang2025sparse}, some of training-free methods retain full attention for the denoising steps close to the noise to preserve generation quality. For tuning-based baselines, we evaluate SLA~\cite{zhang2025sla}, ST-SWA tuned with LoRA and Top-K sparse attention implemented with VMoBA~\cite{wu2025vmoba} mentioned in Sec.~\ref{sec:preliminary}. Both methods are fine-tuned under the same computational budget and follow their original training configurations to ensure a fair comparison. For end-to-end speedup evaluation, we adopt the official open-source kernels released by these methods.

\noindent\textbf{Training Configuration.}
For all tuning-based methods, we apply LoRA to the attention projection layers ($W^Q$, $W^K$, $W^V$, $W^O$) with rank $r=128$ and $\alpha=256$. 
For SALAD, we additionally fine-tune the scale-related modules~(see Fig.~\ref{fig:salad}). 
All models are trained with a batch size of 8. 
For Wan2.1-1.3B, training takes 1,600 steps ($\sim$20 GPU hours) at 480p and 1,000 steps ($\sim$30 GPU hours) at 720p. 
For Wan2.1-14B at 480p, training takes 600 steps ($\sim$20 GPU hours).

\noindent\textbf{Evaluation Metrics.} 
We evaluate the quality of generated videos using the VBench~\cite{huang2024vbench} and Vision Reward~\cite{xu2024visionreward}. 
From VBench, we report \textit{Subject Consistency} (SC), \textit{Background Consistency} (BC), \textit{Image Quality} (IQ), and \textit{Text Consistency} (TC). 
We use the extended prompt set of VBench to generate videos and generate 2 videos for each prompt.
Comparisons with baseline video quality on WAN1.3B and the initialization study are conducted on the full set of VBench prompts.
Besides, WAN14B evaluation, parameter-efficiency comparisons with LoRA and other ablation studies are performed on a subset of VBench prompts.
All ablation studies are conducted on WAN1.3B with 480p resolution.
For efficiency evaluation, we report \textit{Sparsity} (the proportion of computations reduced by sparse attention) and end-to-end \textit{speedup}.
All speedup measurements are carried out on a single GPU with a batch size of 1.

\subsection{Main Results}

\textbf{Comparison with Sparse Attention Methods.} 
We compare SALAD with state-of-the-art sparse attention methods at both 30k and 100k token lengths, as summarized in Tab.~\ref{tab:main_result}. 
In the 30k-token setting, when applied to the spatial-temporal SWA, SALAD consistently outperforms prior approaches across all four evaluation metrics, achieving a $1.72\times$ speedup. Existing post-training sparse attention methods exhibit noticeable quality degradation at 90\% sparsity, even when retaining full attention for a subset of diffusion steps. Although SLA~\cite{zhang2025sla} reduces training cost, it still requires 20,000 video samples and 2,000 training steps with a batch size of 64 to preserve generation quality, and degrades under our limited training budget. LoRA tuning partially restores image fidelity and text alignment; however, the overall performance remains below the dense baseline, even when 90\% sparse attention is applied to only 70\% of diffusion timesteps (63\% overall sparsity). In contrast, SALAD achieves 90\% overall sparsity while maintaining video quality comparable to, and even surpassing, the dense model on VBench. In the 100k-token setting, SALAD achieves a $2.03\times$ speedup while preserving performance comparable to full-attention baselines. 

Notably, SALAD extends existing sparse attention methods and adopts their released CUDA kernels. Consequently, its acceleration depends on kernel efficiency.The Top-K sparse attn is built upon VMoBA~\cite{wu2025vmoba}, whose released kernel is not fully optimized, thereby limiting the observed speedup of Top-K.

We also conduct a preliminary experiment on Wan2.1-14B. As shown in Tab.~2, using ST-SWA with SALAD and post-tuning branch dropping achieves about $1.59\times$ speedup while maintaining performance close to the full-attention model, demonstrating the feasibility of SALAD on large-scale models.

\begin{figure*}[t!]
\centering
\includegraphics[width=0.89\columnwidth]{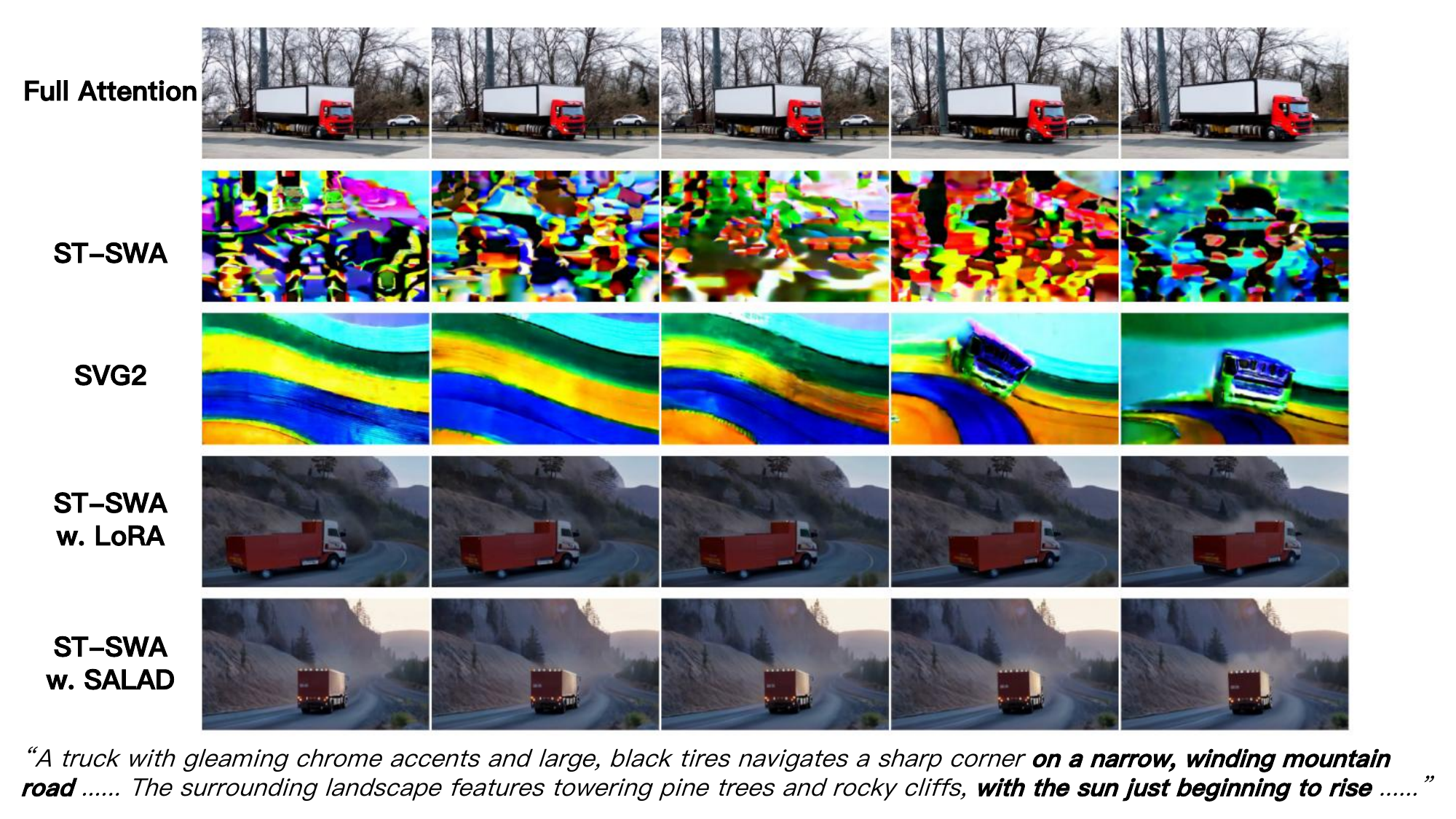}
\caption{\textbf{Generated Video Example of SALAD and other baselines.}}
\label{fig:main_example}
\end{figure*}

\begin{table}[t]
\centering
\caption{\textbf{Quality and efficiency comparison on WAN14B.}}
\label{tab:salad_result_wan14b}
\resizebox{0.85\columnwidth}{!}{
\begin{tabular}{@{} c ccccc cc @{}}
\toprule
\textbf{Method} & SC↑ & BC↑ & IQ↑ & TC↑ & VR↑ & Sparsity↑ & Speedup↑ \\
\midrule
Original (Full Attention) & 94.72 & 97.53 & 62.62 & 27.55 & 0.0938 & 0\% & — \\
\midrule
SVG2 & 93.77 & 97.30 & 58.66 & 24.71 & 0.0601 & 70\% & 1.37$\times$ \\
PAROAtten & 94.01 & 95.82 & 60.02 & 24.78 & 0.0511 & 70\% & 1.20$\times$ \\
\rowcolor{gray!12}
SALAD & \textbf{97.61} & 96.55 & \textbf{65.88} & 27.74 & {0.1243} & 88\% & \textbf{1.52$\times$} \\
\rowcolor{gray!12}
SALAD w. Post-tuning-Drop & {95.61} & \textbf{97.35} & {65.05} & \textbf{28.35} & \textbf{0.1373} & 88\% & \textbf{1.59$\times$} \\
\bottomrule
\end{tabular}
}
\end{table}

\begin{table*}[t]
\centering
\caption{\textbf{Performance and trainable parameters comparison with LoRA.}}
\label{tab:trainable_params}
\renewcommand{\arraystretch}{0.88}
\resizebox{0.85\textwidth}{!}{
\begin{tabular}{@{} l ccccc @{}}
\toprule
\textbf{Method}          & SC~(↑)   & BC~(↑)   & IQ~(↑)   & TC~(↑) &
Trainable Parameters \\
\midrule
Original (Full Attention) & 94.19 & 96.76 & 70.17 & 28.95 & — \\
\midrule
LoRA (r = 128)  & 96.77 & 96.73 & 67.80 & 26.87 & 94M \\
LoRA (r = 256)  & 90.54 & 96.25 & 63.75 & 25.25 & 189M \\
LoRA (r = 512)  & 91.30 & 94.31 & 66.39 & 26.19 & 377M \\
\rowcolor{gray!12}
SALAD           & \textbf{97.21} & \textbf{96.83} & \textbf{69.41} & 25.56 & 165M \\
\bottomrule
\end{tabular}
}
\end{table*}

\noindent\textbf{Comparison with LoRA.} We compared SALAD with various LoRA configurations under 90\% sparsity. As shown in Tab.~\ref{tab:trainable_params}, increasing the number of LoRA parameters during model tuning does not consistently improve video quality, with the best performance achieved at rank 128. SALAD outperforms almost all the LoRA variants in SC, BC, and IQ, while maintaining comparable TC, using fewer trainable parameters than LoRA at rank 256. This suggests that the observed gains in video quality arise from the method's intrinsic design rather than an increase in parameter count.

\subsection{Ablation Study}
\label{sec:ablation}
\textbf{Multi-level Dynamic-Static Scaling.} Tab.~\ref{tab:ablation_arc} presents an ablation study on how each design component affects the video quality. 
Adding the layer-wise static scaling projection alone improves subject consistency and image quality but reduces background and text consistency.
Incorporating the input-timestep-aware dynamic scaling alongside the projection enhances performance across all metrics, indicating that the dynamic scaling is essential for coordinating the two branches' output.
Compared with the Non-shared weight variant (i.e., sparse and linear attn use independent $W^Q$, $W^K$, $W^V$, $W^O$), our design achieves higher scores on all metrics while introducing substantially fewer parameters, demonstrating that SALAD not only optimizes performance but also provides a favorable trade-off between accuracy and parameter efficiency.

\begin{table*}[t!]
\centering
\caption{\textbf{Performance comparison across different architecture designs.} Here, $D$ denotes the number of channels of the input hidden states and Q/K/V. For the parameter count of each projection matrix, $W^{Q,K,V,O}$ is $D \times D$, the static scaling projection layer for the linear attention branch introduces $D \times D$ parameters, and the weight of linear layer in dynamic scaling operation has $D \times 1$ parameters.}
\resizebox{0.85\textwidth}{!}{\begin{tabular}{c c c c c c}
\toprule
Architecture Design & SC~(↑) & BC~(↑) & IQ~(↑) & TC~(↑) & Added Parameters\\
\midrule
Nonshared & 96.02 & 96.07  & 66.87 & 24.92 & $4D\times D$\\
\midrule
Shared   & 95.82 & 96.88  & 68.50 & 23.35 & 0\\
Shared + Static Scaling   & 95.94 & 95.22  & 68.72 & 23.14 & $D \times D$\\
\rowcolor{gray!12}
Shared + Static \& Dynamic Scaling   & 97.21 & 96.83  & 69.41 & 25.56 & $D\times D + D$\\
\bottomrule
\end{tabular}}
\label{tab:ablation_arc}
\end{table*}

\noindent\textbf{Initalization of Static Scaling layers.} We experiment SALAD with random initialization settings when applying static scaling. Zero initialization has lower training loss compared with random initialization. After finetuning, the model with zero initialization projection module demonstrates higher subject, background, and text consistency as shown in Tab.~\ref{tab:init} (a). 

\noindent\textbf{Input-Timestep-Aware Scaling.} 
To further investigate how the dynamic scaling facilitates performance recovery, we examine two variants. \textit{Constant Scale} variant replaces the dynamic scaler with a fixed scalar $\lambda = 0.5$.
The \textit{Detached} variant blocks the gradient flow from the Dynamic Scaler to the input features while keeping the Scaler’s internal channel-mixing linear layers trainable. 
By truncating gradient flow to the input, this variant helps determine if backpropagating gradients to the backbone provides any additional benefit beyond the dynamic scaling itself.

As shown in Tab.~\ref{tab:gate_design} (b), \textit{Constant Scale} improves video quality, indicating that adjusting the contribution of the linear attention during both training and inference enhances recovery. However, it performs worse than the dynamic scaling values, suggesting that adaptive scaling more effectively regulates the auxiliary role of the linear attention branch. \textit{Detached variant} also achieves higher video quality, implying that the improvement arises primarily from dynamic scaling of the linear attention rather than from gradient flow to the input.

\begin{table}[t!]
\centering
\small
\caption{Ablation study on scaling related modules. (↑) indicates higher is better.}
\begin{minipage}{0.4\textwidth}
\centering
(a) Initialization
\label{tab:init}
\begin{tabular}{ccccc}
\toprule
Initialization & SC~(↑) & BC~(↑) & IQ~(↑) & TC~(↑) \\
\midrule
Random         & 96.16  & 95.51  & 65.67  & 25.70  \\
\rowcolor{gray!12}
Zero           & \textbf{96.54} & \textbf{96.37} & \textbf{66.09} & 25.55  \\
\bottomrule
\end{tabular}
\end{minipage}%
\hfill
\begin{minipage}{0.49\textwidth}
\centering
(b) Dynamic Scaling Design
\label{tab:gate_design}
\begin{tabular}{ccccc}
\toprule
Methods    & SC~(↑) & BC~(↑) & IQ~(↑) & TC~(↑) \\
\midrule
Static Only       & 95.94  & 95.22  & 68.72  & 23.14  \\
\rowcolor{gray!12}
+ Dynamic        & 97.21  & 96.83  & 69.41  & 25.56  \\
+ Constant & 97.45  & 96.55  & 64.96  & 24.23  \\
\rowcolor{gray!12}
+ Detached     & 97.00  & 96.60  & 72.90  & 25.89  \\
\bottomrule
\end{tabular}
\end{minipage}

\label{tab:ablation-init-gate}
\end{table}



\subsection{Additional Observations}
\label{sec:add_observe}




\noindent\textbf{Effect of the Branch.}
To analyze the role of the linear branch, we disable it at inference after SALAD training (denoted as SALAD-drop) and compare the results with full SALAD and LoRA-only tuning. 
As shown in Fig.~\ref{fig:branch_analysis}, LoRA-only tuning fails to recover long-range dependencies under high sparsity, leading to missing prompt objects (e.g., fishing boats, top) and subject duplication (e.g., two dogs, bottom). 
SALAD-drop improves video–text consistency but exhibits noticeable detail loss (e.g., the boat in sample 1 and the dog’s tail in sample 2). 
In contrast, the full SALAD model produces richer details and better temporal continuity. 
These results indicate that the sparse attention primarily restores core generation quality, while the linear branch provides complementary token interactions to enhance fine details and coherence.

\section{Conclusion}
We present \textbf{SALAD}, an efficient sparse-linear attention for Diffusion Transformers. 
To alleviate the cross-token information loss caused by ultra-sparse attention, we introduce a lightweight linear attention branch in parallel with sparse attention. 
We observe that the linear attn captures low-rank yet critical global information missed by sparse attention, while sparse attention models the bulk of the sequence. 
To regulate this balance, a multi-level static-dynamic scaling strategy is introduced. 
Across models and sequence lengths, SALAD achieves up to \textbf{90\% sparsity} and \textbf{1.52-2.03$\times$ inference speedup} while maintaining generation quality comparable to full attention. 

\section{Acknowledge}
\label{sec:acknowledge}
This work was supported by National Natural Science Foundation of China (No. 62506197, 62325405,
62104128, U19B2019, U21B2031, 61832007, 62204164), Tsinghua EE Xilinx AI Research Fund, Beijing National Research Center for Information Science and Technology (BNRist) and Kuaishou Technology.


\clearpage  


%
%
\bibliographystyle{splncs04}
\bibliography{main}

\clearpage
\appendix
\section{Appendix}
\subsection{Sparse Attention Implementation Details}
\label{app:sparse_details}
\subsubsection{Spatial-Temporal Sliding Window Attention}
As mentioned in the Sec.~\ref{sec:preliminary}, we now provide more details about the Spatial-Temporal Sliding Window Attention (ST-SWA).

Previous studies~\cite{xi2025sparse} have shown that attention heads in Video Diffusion Transformers exhibit two characteristic sparse patterns: \textit{spatial local attention}, where a token primarily attends to tokens within the same frame, and \textit{temporal attention}, where a token focuses on tokens at the same spatial location across neighboring frames. In these models, 3D video latents of size $[H_h, W_h, F_h]$ are flattened into a 1D token sequence.

A common flattening strategy treats the temporal dimension as the slowest-varying index, ensuring that spatially adjacent tokens within each frame remain consecutive. In contrast, tokens at the same spatial location across frames are separated by $H_h \times W_h$ positions, causing high temporal attention scores to form a multi-diagonal pattern in the attention map. As a result, spatial locality can be effectively captured using local sliding-window attention, whereas temporal locality cannot be modeled well under this default ordering.

To address this, for attention heads that exhibit temporal locality, we apply a token permutation that groups tokens by spatial location, making temporal neighbors consecutive in the flattened sequence. This permutation realigns temporal locality with the diagonal structure of the attention map. After this, a sliding\text{-}window attention pass with an appropriately chosen window size can directly and effectively capture local temporal dependencies.

In our implementation of spatial-temporal attention, we adopt the headwise permutation kernel from SVG~\cite{xi2025sparse} and the headwise sliding-window attention kernel from DiTFastAttnV2~\cite{zhang2025ditfastattnv2}. For configuring the window size of each attention head, we follow the procedure of DiTFastAttnV2 and use eight VBench video prompts as the profiling dataset. We then greedily select the smallest window size whose induced error does not exceed a pre-defined threshold. The error threshold $\delta$ (maximum RSE after applying the sparse attention) is set to $2.0$.


\subsubsection{Top-K Block Sparse Attention.}
Instead of applying a fixing pattern, Top-K, or the extension Top-P, dynamically select the key and value tokens that are more likely to have high attention scores. The calculation of Top-K block sparse attention includes three steps. \\
\textbf{Step 1}: The input tokens is first partitioned in to blocks and the mean of each block is calculated. \\
\textbf{Step 2}: The similarity of query tokens and the mean of key blocks is calculated and the top k blocks are selected. \\
\textbf{Step 3}: each query token only performs attention with the selected key blocks. \\
For the Top-K sparse attention implementation, we adopt the Top-K mode from VMoBA's implementation. In this work, we use k=8 for LoRA tuning and k=4 for SALAD tuning. 

\begin{figure}[t!]  
  \centering
  \includegraphics[width=0.8\columnwidth]{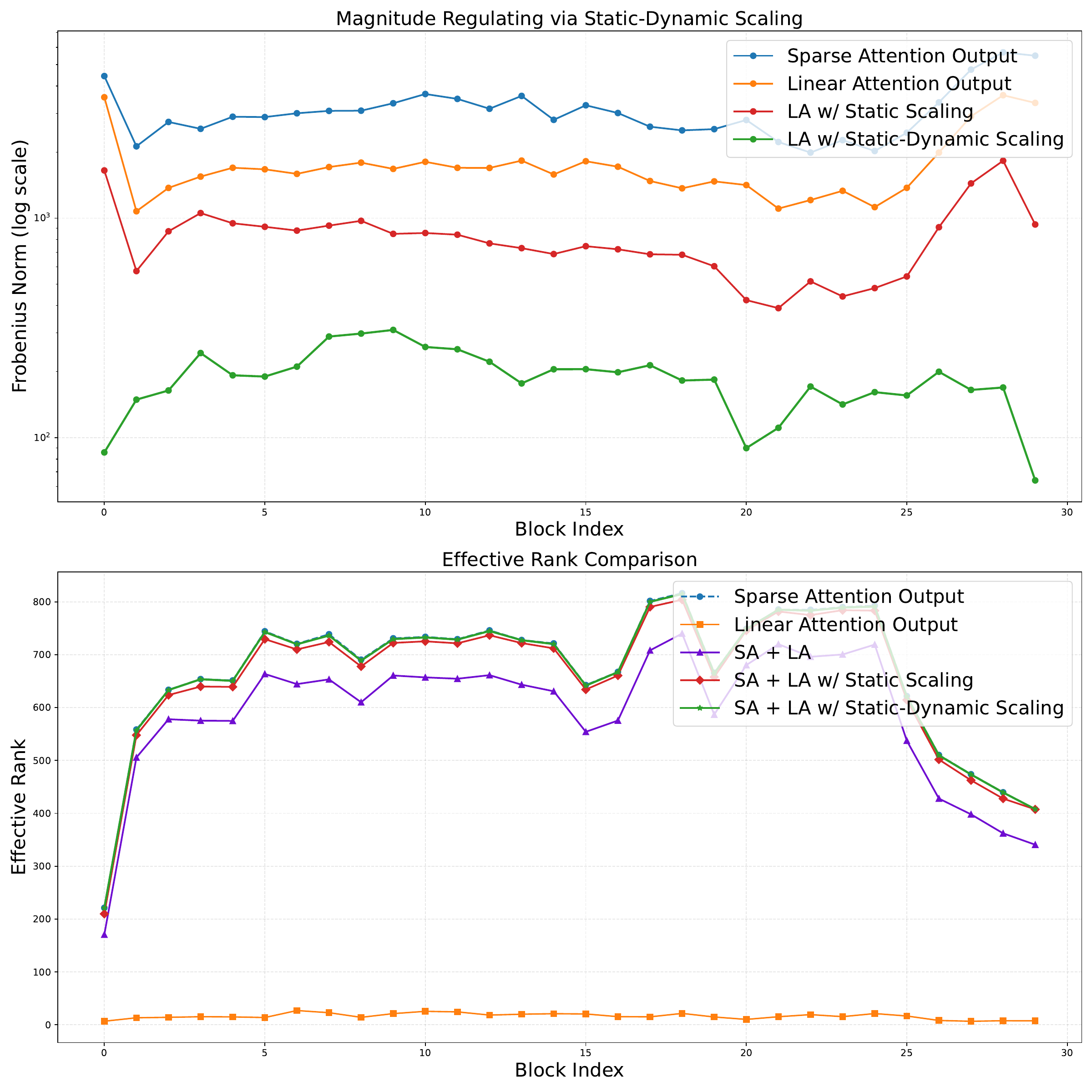}
    \caption{\textbf{Impact of Multi-level Static-Dynamic Scaling on linear attention Regulation and Rank Restoring.} Both metrics are evaluated on the attention outputs of a Wan2.1-1.3B model at sampling timestep $t=20$.
    \textbf{(Top) The Frobenius norms of attention outputs across different Transformer blocks.} The results demonstrate that our multi-level scaling strategy effectively down-weights the linear attention output magnitude to a much smaller range relative to the sparse branch. 
    \textbf{(Bottom) The effective rank ($r_\mathrm{eff}$) across blocks.} While naive summation leads to a rank erosion, introducing our scaling strategy progressively restores the joint rank. Finally, the green curve (sparse + linear attention w/ static-dynamic scaling) matches the effective rank of the blue curve (pure sparse attention).}
  \label{fig:value_rank}
\end{figure}

\subsection{Detailed Analysis of Multi-level Static-Dynamic Scaling}
\label{app:rank_analysis}

In Sec.~\ref{sec:linear_for_sparse}, we identified that there existing \textit{rank imbalance} and \textit{rank degradation}~(see the Fig.~\ref{fig:rank_and_branch_effect}) in naive sparse-linear attention. To address these, we introduced a multi-level static-dynamic scaling strategy in Sec.~\ref{sec:multi_level_scaling}. In this section, we provide a structured analysis to elucidate how this strategy enables effective sparse and linear attention branches fusion.

Note that, in our analysis, we measure the representation rank using the \textit{effective rank} metric, which is defined based on the entropy of normalized singular values. 
Given the singular values $\{\sigma_i\}$ of a matrix, we compute 
\begin{equation}
p_i = \frac{\sigma_i}{\sum_j \sigma_j},
\end{equation}
and define the effective rank as
\begin{equation}
\label{eq:effective_rank}
r_{\mathrm{eff}} = \exp\Bigg(- \sum_i p_i \log p_i \Bigg).
\end{equation}
This metric captures the intrinsic dimensionality of the representation and provides a stable estimate of rank under noisy or approximately low-rank conditions.

\subsubsection{Linear Attention Regulation and Rank Restoration}

To regulate the contribution of the linear attention branch, our scaling strategy explicitly controls its numerical magnitude before aggregation. 
To visualize this effect, we measure the Frobenius norm of the outputs of the linear attention branch, defined as $\|\mathbf{X}\|_F=\sqrt{\sum_{i,j}X_{ij}^2}$ for a matrix $\mathbf{X}$.

As illustrated in Fig.~\ref{fig:value_rank}~(Top), the naive linear attention output often exhibits a Frobenius norm close to that of the sparse branch, indicating that the linear branch contributes a comparable magnitude to the final representation. 
However, as shown in Fig.~\ref{fig:value_rank}~(Bottom), the two branches exhibit a severe rank imbalance, where the linear attention output is significantly lower-rank than the sparse branch. 
Directly summing these two branches therefore leads to rank erosion, resulting in a combined representation whose effective rank is even lower than that of the sparse attention alone.

With the introduction of our scaling strategy, the magnitude of the linear attention branch is progressively suppressed. 
As shown in Fig.~\ref{fig:value_rank}~(Top), applying the static projection already reduces the energy of the linear branch, while further incorporating the input-timestep-aware dynamic scaling suppresses it even more. 
Correspondingly, the effective rank of the fused representation gradually recovers, as illustrated in Fig.~\ref{fig:value_rank}~(Bottom).

Notably, although the layer-wise static projection does not explicitly impose a down-weighting constraint during training, it nevertheless learns to attenuate the contribution of the linear attention branch, implicitly indicating the necessity of regulating linear attention. 
Furthermore, the dynamic scaling mechanism further reduces the magnitude of the linear branch outputs, validating its complementary role in controlling their influence.

Overall, as the contribution of the linear branch is progressively regulated, the rank of the combined representation increases accordingly, demonstrating that the proposed scaling strategy effectively mitigates the adverse impact of the low-rank linear branch on the sparse attention outputs.

\begin{figure}[t!]  
  \centering
  \includegraphics[width=0.8\columnwidth]{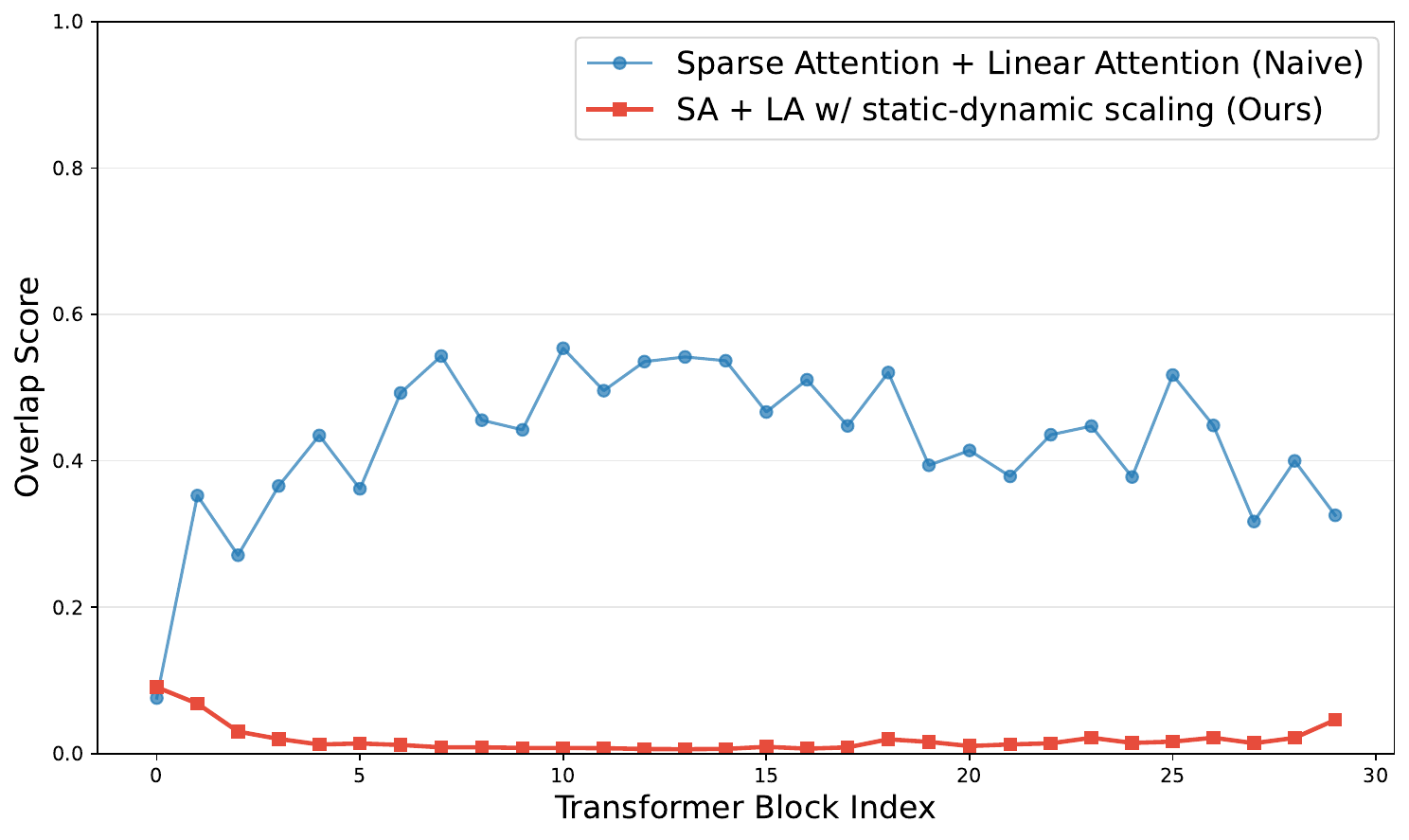}
    \caption{\textbf{PCA subspace overlap.} We collect the output of sparse and linear attention at diffusion timestep 20 across different blocks in WAN1.3B and compute the overlap score in PCA subspace between them.}
  \label{fig:pca_subspace}
\end{figure}

\begin{figure}[t!]  
  \centering
  \includegraphics[width=0.9\columnwidth]{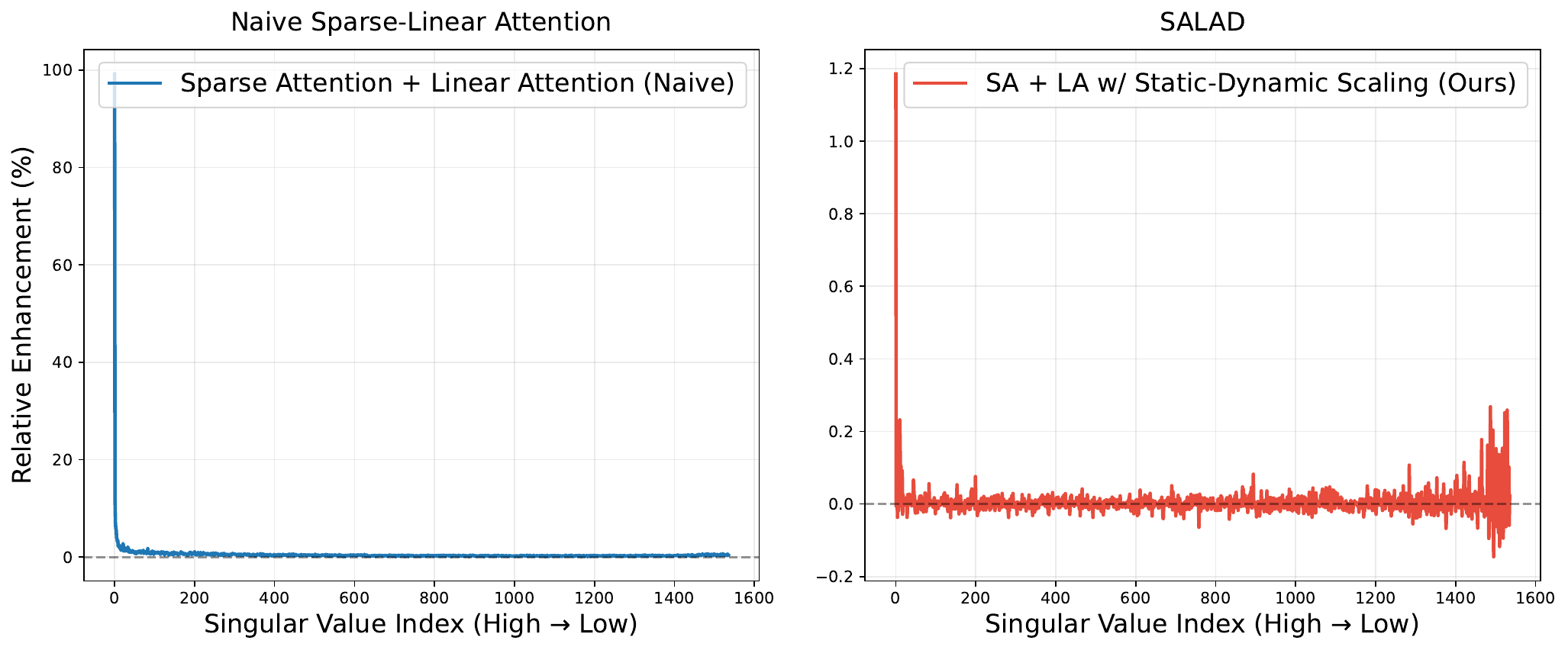}
    \caption{\textbf{Relative Singular Value Enhancement.}}
  \label{fig:relative_enhancing}
\end{figure}

\subsubsection{Subspace Decoupling and Spectral Re-attribution}
Beyond balancing the contribution of sparse and linear attention branches, we further investigate scaling strategy can help the two branch learn different information.

\textbf{Subspace Overlap Analysis.}
To quantify the relationship between the sparse and linear branches, we measure the overlap between their principal subspaces.
Specifically, given the output matrices produced by the sparse attention branch $\mathbf{A}$ and the linear attention branch $\mathbf{B}$, we compute their top-$k$ principal components via PCA.
Let $\mathbf{V}_A$ and $\mathbf{V}_B$ denote the corresponding orthonormal bases.
We then compute the overlap score
\[
\text{Overlap}(\mathbf{A}, \mathbf{B}) =
\frac{1}{k}\|\mathbf{V}_A^\top \mathbf{V}_B\|_F^2 ,
\]
which measures the average squared cosine similarity between the two $k$-dimensional subspaces.

As shown in Fig.~\ref{fig:pca_subspace}, introducing the proposed scaling strategy reduces the overlap score of the two branches.
This indicates that the linear branch becomes less aligned with the dominant subspace of the sparse branch.
In other words, the scaling mechanism encourages the two branches to encode complementary information rather than redundantly modeling the same dominant directions.

\textbf{Spectral Tail Revitalization.}
To further analyze how the linear branch affects the representation spectrum, we perform a singular value decomposition (SVD) analysis on the attention outputs.
Let $\sigma_i(\cdot)$ denote the $i$-th singular value.
We measure the relative singular value change after introducing the linear branch:
\[
\text{RelSV}_i =
\frac{\sigma_i(\mathbf{A} + \lambda \mathbf{B}) - \sigma_i(\mathbf{A})}
{\sigma_i(\mathbf{A})}.
\]

Interestingly, although the overall effective rank of the representation remains comparable to that of the sparse attention baseline, the singular value distribution changes noticeably after introducing the scaled linear branch.
As illustrated in Fig.~\ref{fig:relative_enhancing}~(Right), the leading singular values and the tail-end singular values both exhibit a increase compared with the sparse attention output.
This suggests that the linear branch can enriches the weaker spectral components, encouraging the sparse attention backbone to model additional information that was previously underrepresented.

In contrast, when the two branches are combined via naive summation (without scaling), Fig.~\ref{fig:relative_enhancing}~(Left) reveals that the dominant singular value of the sparse attention increases dramatically, far more than the rest of the spectrum.
This indicates that the linear branch primarily reinforces directions already captured by sparse attention, leading to redundant representations rather than complementary information.
Therefore, the proposed static-dynamic scaling strategy plays a crucial role in preventing spectral dominance and enabling the linear branch to act as a complementary refinement to the sparse backbone.

\subsubsection{Theoretical Interpretation via Weyl's Inequality}

The effectiveness of our scaling strategy in mitigating rank erosion can be qualitatively understood through \textit{Weyl's inequality}. 
For a matrix $\mathbf{A}$ perturbed by another matrix $\mathbf{B}$, the change of each singular value is bounded by the spectral norm (largest singular value) of the perturbation:
\begin{equation}
    |\sigma_i(\mathbf{A}+\mathbf{B})-\sigma_i(\mathbf{A})| \le \|\mathbf{B}\|_2 .
\end{equation}

This inequality implies that the magnitude of the perturbation $\mathbf{B}$ directly controls how strongly the singular value spectrum of $\mathbf{A}$ can be altered. 
In the naive summation setting, the linear branch may have a relatively large norm, allowing it to significantly distort the spectrum of the sparse attention output. 
In particular, unfavorable interactions between the two branches can suppress weaker singular directions, leading to the observed rank erosion.

Our scaling strategy effectively reduces the overall magnitude of the linear branch before aggregation, thereby tightening the Weyl bound and limiting the possible perturbation to the singular values of $\mathbf{A}$. 
As a result, the dominant spectral structure of the sparse attention branch is preserved while allowing the linear branch to inject complementary information in less dominant directions.

Our hierarchical scaling transforms the integration into a constrained perturbation: $|\sigma_i(\mathbf{A} + \lambda \mathbf{B}) - \sigma_i(\mathbf{A})| \le \lambda \|\mathbf{B}\|_2$. By enforcing $\lambda \ll 1$, we ensure the perturbation remains small relative to the significant singular values of $\mathbf{A}$. This prevents spectral collapse while allowing the linear branch to provide complementary information.

\subsection{Details of Post-training Branch Dropping}
\label{app:branch_drop}


In Sec.~\ref{sec:post_branch_drop}, we mentioned that some linear attention branches can be dropped to gain more efficiency while maintaining performance. We now provide more details.

We first analyze the distribution of the input-timestep-aware dynamic scaling values across denoising timesteps. 
The scaling value reflects the relative contribution of the linear attention branch under a specific input prompt and timestep. 
Since one scaling factor is assigned to each block, it also implicitly captures the block-wise demand for the linear attention branch.
Importantly, these scaling values can be computed with negligible overhead during inference, making them a natural indicator for deciding whether a linear attention branch should be executed or skipped. 
Therefore, we introduce a threshold-based rule: if the scaling value of a branch falls below a threshold, the corresponding linear attention branch is dropped to reduce computation.

\begin{figure}[t]  
  \centering
  \includegraphics[width=0.8\columnwidth]{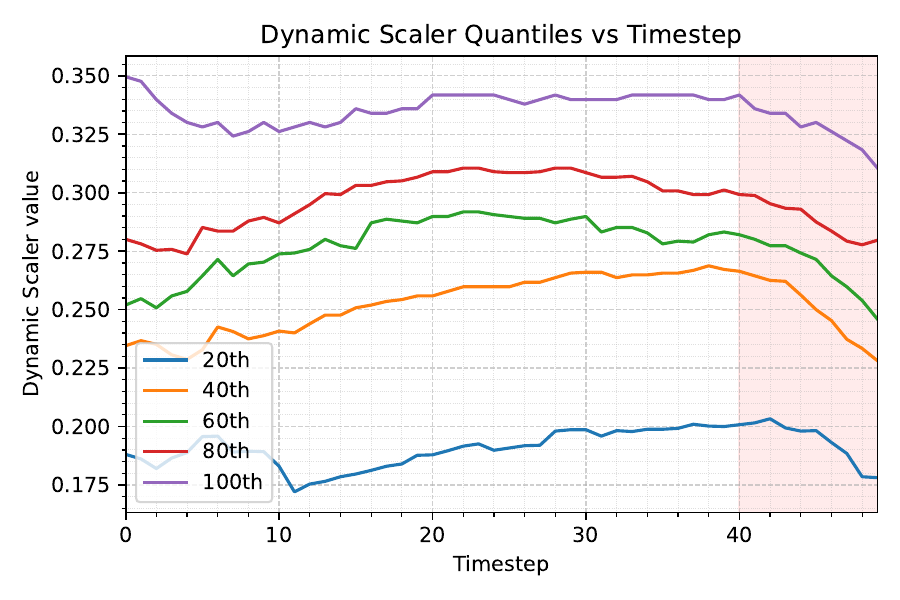}
    \caption{\textbf{Quantile trends of dynamic scaling values across diffusion timesteps.}
    We plot several quantiles of the input-timestep-aware scaling values over the denoising process. 
    These quantiles reflect the overall magnitude of scaling factors. 
    The consistent downward shift at later timesteps indicates that the contribution of the linear attention branch generally diminishes as the denoising approaches the final image..}
  \label{fig:quantile}
\end{figure}

\begin{figure}[t]  
  \centering
  \includegraphics[width=0.8\columnwidth]{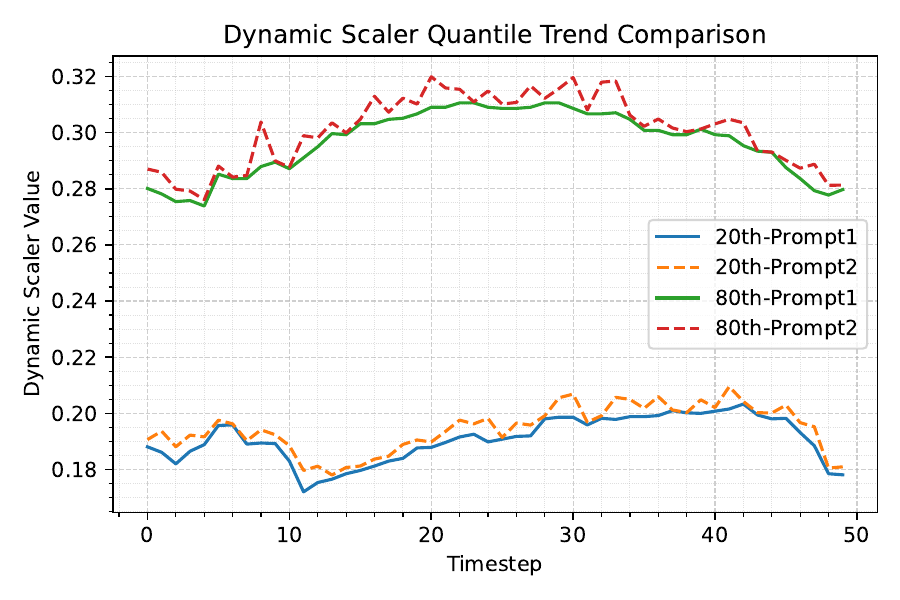}
  \caption{\textbf{Comparison of dynamic scaling values quantiles across timesteps for two different prompts. }The quantile curves exhibit highly similar trends between prompts, indicating that the distribution of dynamic scaling values generalizes well across prompts.}
  \label{fig:quantile_consi}
\end{figure}

As illustrated in Fig.~\ref{fig:quantile}, the overall magnitude of scaling values varies across different diffusion timesteps. 
In particular, the 20th, 40th, 60th, and 80th percentiles consistently shift toward smaller values at later timesteps. 
In Fig.~\ref{fig:quantile_consi}, we further observe that for different input prompts, the quantile statistics at the same timestep are highly consistent, indicating that the distribution of scaling values is largely timestep-dependent rather than prompt-dependent.

Motivated by this observation, we adopt a shared threshold for all blocks at the same timestep. 
Specifically, for a model with $N$ blocks, dropping the linear attention branches of $M$ blocks corresponds to a drop ratio of $M/N$. 
By controlling this ratio through a timestep-wise threshold $\tau_t$, we can regulate the fraction of linear branches that are skipped at each timestep. 
\textbf{Since the scaling statistics are stable across prompts, the thresholds $\tau_t$ can be determined using a small calibration set}.

We explore two strategies for determining $\tau_t$. 
First, in a \textit{coarse-grained strategy}, we determine a threshold $\tau_0$ at timestep $t=0$ (the input latent is pure noise) such that a desired fraction of linear branches is dropped. 
We then apply the same threshold to all timesteps: 
$\tau_t=\tau_0$ for $t=0,\dots,T-1$, where T is the number of diffusion sampling timesteps.

Because the scaling values naturally vary across timesteps, this constant threshold automatically results in different dropping ratios at different stages of the denoising process, thereby adapting to the varying demand for linear attention.

Second, in a \textit{fine-grained strategy}, we manually specify the desired dropping ratio for several timestep intervals and calibrate the corresponding thresholds $\tau_t$ for each interval using the same calibration set. 
This provides finer control over the dropping behavior while still requiring only lightweight calibration.

We conduct an ablation study on the two dropping schemes using a subset of VBench, as shown in Tab.~\ref{tab:gate_drop}. 
We find that even the simple coarse-grained strategy achieves performance comparable to SALAD without any dropping, while providing a slight efficiency gain. 

In Tab.~\ref{tab:salad_result_wan14b}, we also adopt the coarse variant as the drop strategy for SALAD w/ Post-tuning-Drop, achieving an average branch dropping ratio of about $50\%$ across different timesteps.

For comparison, we also evaluate a random dropping strategy that removes the same proportion of linear attention branches. The performance of this baseline is noticeably worse, further demonstrating that the input-timestep-aware dynamic scaling values provide a reliable indicator for determining which linear branches can be safely pruned.

\begin{table}[h]
  \centering
  \caption{Qualitative results with different drop strategies}
  \label{tab:gate_drop}
  \begin{tabular}{lcccc|c}
    \hline
    Drop Strategy & SC~(↑) & BC~(↑) & IQ~(↑) & TC~(↑) & Overall Latency of inference(s)\\
    \hline
    w.o. Drop       & 97.21  & \textbf{96.83} & 69.41  & \textbf{25.56} & 98.9\\
    fine-grained $^{*}$         & 97.32  & 95.84  & 69.27  & 25.08 & 96.6\\
    coarse-grained $^{\dagger}$     & \textbf{97.44}  & 95.78  & \textbf{69.76}  & 24.66 & 97.4\\
    Random$^{\ddagger}$  & 96.89  & 96.27  & 66.79  & 24.40 & 96.8\\
    \hline
  \end{tabular}
  \vspace{2pt}

 {\footnotesize
  \raggedright
  $^{*}$\textbf{fine-grained} drops $20\%$ of branches during the first $40$ timesteps and $30\%$ during the last $10$ timesteps. \\
  $^{\dagger}$\textbf{coarse-grained} removes branches whose dynamic scaling values fall below the lowest $20\%$ at the first timestep, using this threshold uniformly across all timesteps. \\
  $^{\ddagger}$\textbf{Random} drops each branch independently with a probability of $20\%$. \\  }
\end{table}

\subsection{Ablation on the Non-Linear Function in the Input-Timestep-Aware Scaling Module}
\label{app:non-linear}

In this section, we investigate the choice of the non-linear function used in the Dynamic Scaler. As illustrated in the main paper Fig.~\ref{fig:salad}, we introduce a non-linearity to increase the expressive capacity of the Input-Timestep-Aware Scaling Module, and we adopt the sigmoid function since a gating range constrained to $[0,1]$ empirically improves the performance of SALAD. Tab.~\ref{tab:non-linear} reports the ablation results across different non-linear functions. Sigmoid yields the best performance, while ReLU and Tanh underperform. This suggests that restricting the dynamic scaling values to $[0,1]$ is sufficient and beneficial, which is also consistent with our visualization results.


\begin{table}[h]
  \centering
  \caption{Comparison of Non-Linear Functions in the Dynamic Scaler}
  \label{tab:non-linear}
  \begin{tabular}{lcccc}
    \hline
    Non-Linear Function & SC~(↑) & BC~(↑) & IQ~(↑) & TC~(↑) \\
    \hline
    Tanh         & 96.89 & 96.25 & 66.20 & 26.97 \\
    ReLU         & 96.77 & 96.10 & 67.12 & \textbf{28.26} \\
    \rowcolor{gray!12}
    Sigmoid      & \textbf{97.21} & \textbf{96.83} & \textbf{69.41} & 25.56 \\
    \hline
  \end{tabular}
\end{table}

\begin{figure*}[t!]
\centering
\includegraphics[width=0.8\columnwidth]{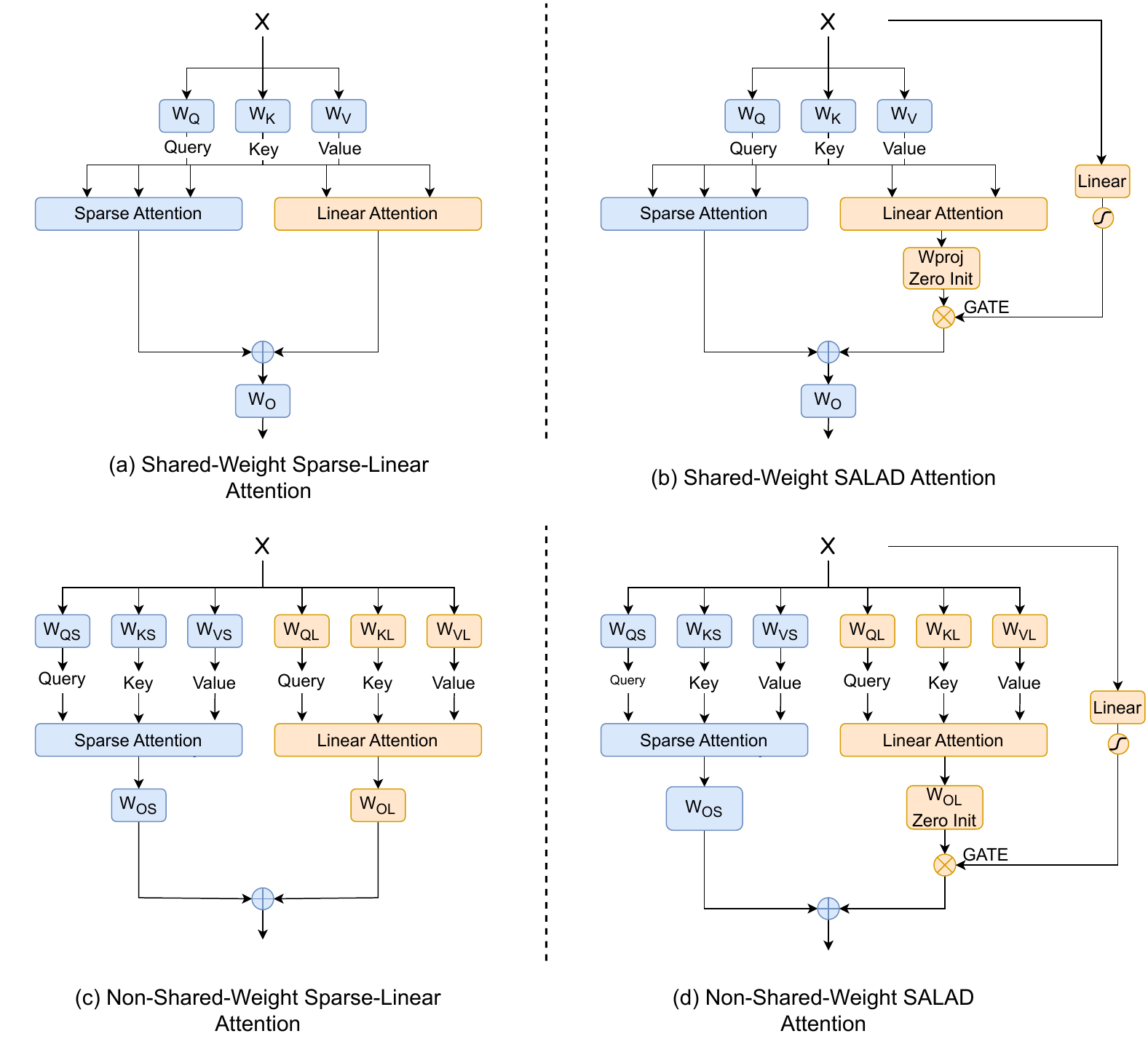}
\caption{\textbf{Shared-Weight and Non-Shared-Weight Architecture of Naive Sparse-Linear attention and SALAD Attention.}}
\label{fig:shared-nonshared}
\end{figure*}

\subsection{Details of Non-Shared Sparse-Linear Attention}
\label{app:non-shared}

As discussed in Sec.~\ref{sec:ablation} and Tab.~\ref{tab:ablation_arc}, the non--shared-weight sparse--linear attention architecture maintains two independent parameter sets for the sparse and linear attention branches, resulting in lower parameter efficiency compared with the shared-weight design. Although it introduces additional parameters, the non--shared-weight variant can also be adapted to the SALAD formulation. Fig.~\ref{fig:shared-nonshared} illustrates the naive sparse--linear attention (S-L Attn) and the SALAD attention under both shared-weight and non--shared-weight settings.

\begin{table}[t!]
\centering
\caption{\textbf{Performance comparison across shared and non-shared weight SALAD.}}
\begin{tabular}{lcccc}
\toprule
Architecture & SC~(↑) & BC~(↑) & IQ~(↑) & TC~(↑) \\
\midrule
Shared S-L Attn & 95.82 & 96.88  & 68.50 & 23.35 \\
\rowcolor{gray!12}
Shared SALAD & 97.21 & 96.83  & 69.41 & 25.56 \\
NonShared S-L Attn & 96.02 & 96.07  & 66.87 & 24.92 \\
\rowcolor{gray!12}
NonShared SALAD & 97.29 & 96.68  & 68.38 & 27.05 \\
\bottomrule
\end{tabular}
\label{tab:gate_design}
\end{table}

Tab.~\ref{tab:ablation_arc} primarily compares the shared weight SALAD attention against the naive non-shared weight sparse-linear attention to demonstrate the parameter efficiency of our design. To further examine the impact of our approach, we additionally evaluate the non-shared weight SALAD architecture.

For the experiments in Tab.~\ref{fig:shared-nonshared}, we follow the same training configuration for both shared and non-shared variants and evaluate all models on the same VBench subset. Though has more weights, the performance of naive non-shared sparse-linear attention does not significantly increase when compared with the shared sparse-linear attention variant. After incorporating the input-dependent Dynamic Scaler and the zero-initialization strategy, its performance on all the metrics improves substantially.

Meanwhile, the shared-weight and non-shared-weight versions of SALAD achieve comparable overall performance. The non--non-shared-weight variant yields slightly higher SC and TC, whereas the shared-weight model performs better on BC and IQ. These results highlight the parameter efficiency of the shared-weight formulation. To avoid additional memory and latency overhead, we therefore adopt the shared-weight SALAD architecture as our final design.





\subsection{Extended Comparison with LoRA}
\label{app:lora_exp}
In addition to the comparison presented in the main paper, we provide a more detailed evaluation of LoRA under different sparsity strategies in Tab.~\ref{tab:trainable_params}. Specifically, we report additional results for LoRA with multiple ranks under the ST-SWA and TopK ($k=4$) sparsity strategy.

Consistent with the observations in the main text, increasing the LoRA rank does not consistently improve generation quality. The best-performing LoRA configuration generally occurs at moderate ranks (e.g., $r=128$), while larger ranks can even degrade performance. In contrast, SALAD achieves competitive or better performance across most metrics, demonstrating that the improvements primarily arise from the design of the linear attention branch rather than simply increasing the number of tunable parameters.

Note that the trainable parameters reported in Tab.~\ref{tab:trainable_params} include the parameters introduced by LoRA and linear attention branch during training. However, LoRA weights are merged into the base model after training and therefore do not introduce additional parameters during inference. Therefore, the additional parameter overhead reported in the contributions (4.99\%) refers to the persistent parameters required by SALAD during inference.

\begin{table*}[t!]
\centering
\caption{Performance and trainable parameters comparison between SALAD and LoRA.}
\begin{tabular}{p{3cm}|*{5}{c}}
\hline
\multicolumn{1}{c|}{\textbf{Method}} & 
\textbf{Subject} & 
\textbf{Background} & 
\textbf{Image} & 
\textbf{Text} & 
\textbf{Trainable} \\
 & 
{\textbf{Consistency}} & 
{\textbf{Consistency}} & 
{\textbf{Quality}} & 
{\textbf{Consistency}} & 
{\textbf{Parameters}} \\
\hline
Original (Full Attention) & 94.19 & 96.76 & 70.17 & 28.95 & - \\
\hline
ST-SWA \\
\hline
LoRA (r = 64)             & 96.89 & 96.18 & 65.65 & 27.60 & 47M \\
LoRA (r = 128)            & 96.77 & 96.73 & 67.80 & 26.87 & 94M \\
LoRA (r = 256)            & 90.54 & 96.25 & 63.75 & 25.25 & 189M \\
LoRA (r = 512)            & 91.30 & 94.31 & 66.39 & 26.19 & 377M \\
SALAD & 97.21 & 96.83 & 69.41 & 25.56 & 165M \\
\hline
TopK (k=4) \\
\hline
LoRA (r = 128)            & 95.37 & 96.72 & 67.30 & 27.33 & 94M \\
LoRA (r = 256)            & 98.35 & 96.88 & 69.99 & 26.88 & 189M \\
LoRA (r = 512)            & 96.52 & 96.52 & 69.56 & 24.27 & 377M \\
SALAD & 98.16 & 95.93 & 70.33 & 27.61 & 165M \\
\hline
\end{tabular}
\label{tab:trainable_params}
\end{table*}

\subsection{Training Details}
\label{app:training}
\subsubsection{Training Settings}
All training configurations are summarized in Tab.~\ref{tab:training_details}.
We conduct experiments on the Wan2.1 family of video diffusion transformers~\cite{wan2025}. 
Unless otherwise specified, experiments are performed on Wan2.1-1.3B at a resolution of $480 \times 832$ with 77 frames (approximately 30k tokens). 
We also extend Wan2.1-1.3B to a higher resolution setting of 720p (about 100k tokens) and apply SALAD to Wan2.1-14B at 480p.

All training-based methods are fine-tuned using 2,000 videos sampled from the Mixkit dataset under the OpenSora training protocol~\cite{lin2024open}. 
We apply LoRA to the attention projection layers ($W^Q$, $W^K$, $W^V$, $W^O$) with rank $r=128$ and $\alpha=256$. 
For SALAD, we additionally fine-tune the scaling-related modules (see Fig.~\ref{fig:salad}). 

Training is performed on 4 GPUs with a total batch size of 8 and a learning rate of $1\mathrm{e}{-4}$. 
For Wan2.1-1.3B at 480p, training runs for 1,600 steps. 
For the higher-resolution 720p setting, we train for 1,000 steps. 
For Wan2.1-14B at 480p, training requires 600 steps.


\begin{table}[t!]
\centering
\caption{Training details for SALAD fine-tuning under different settings. All experiments use 2,000 videos from the Mixkit dataset under the OpenSora training protocol~\cite{lin2024open}, LoRA on attention projection layers ($W^Q$, $W^K$, $W^V$, $W^O$) with $r=128$, $\alpha=256$, and additionally tune scaling-related modules for SALAD (see Fig.~\ref{fig:salad}). Training is performed on 4 GPUs with total batch size 8, AdamW optimizer, and learning rate $1\mathrm{e}{-4}$.}
\label{tab:training_details}
\small
\begin{tabular}{lccc}
\hline
 & Wan2.1-1.3B (480p) & Wan2.1-1.3B (720p) & Wan2.1-14B (480p) \\
\hline
Model & Wan2.1-1.3B & Wan2.1-1.3B & Wan2.1-14B \\
Resolution & $480 \times 832$ & $720 \times 1280$ & $480 \times 832$ \\
Approx. Tokens & $\sim$30k & $\sim$100k & $\sim$30k \\
Tuning Steps & 1,600 & 1,000 & 600 \\
GPU Hours (est.) & $\sim$20 & $\sim$30 & $\sim$20 \\
Optimizer & AdamW & AdamW & AdamW \\
Learning rate & $1\mathrm{e}{-4}$ & $1\mathrm{e}{-4}$ & $1\mathrm{e}{-4}$ \\
LoRA rank $r$ / $\alpha$ & 128 / 256 & 128 / 256 & 128 / 256 \\
Batch size & 8 (4 GPUs) & 8 (4 GPUs) & 8 (4 GPUs) \\
Training samples & 2,000 (Mixkit) & 2,000 (Mixkit) & 2,000 (Mixkit) \\
\hline
\end{tabular}
\end{table}

\subsubsection{Training Cost}

SALAD requires only 2,000 training videos and a relatively small number of optimization steps. 
For Wan2.1-1.3B at 480p, the entire tuning process takes approximately 20 GPU hours with a batch size of 8. 
Even at higher resolution (720p), the tuning cost remains modest at around 30 GPU hours. 
For the larger Wan2.1-14B model at 480p, SALAD requires only 600 training steps (about 20 GPU hours).

Overall, SALAD enables efficient adaptation of sparse-linear hybrid attention in Video Diffusion Transformers with a substantially lower computational cost than approaches that pretrain sparse-attention architectures from scratch.

For reference, VSA~\cite{zhang2025vsa} is pretrained on 80{,}000 videos using 32 NVIDIA H100 GPUs, while VMoBA~\cite{wu2025vmoba} leverages the Koala-36M dataset (36 million video clips) and 104 NVIDIA H800 GPUs under the token sequence length of 33K. SLA~\cite{zhang2025sla}, a recent linear-attention-based approach for high-sparsity adaptation, is tuned on 20{,}000 self-collected videos—ten times more than SALAD. SLA also requires a batch size of 64 for tuning. Under our experimental setting, SLA fails to recover the performance of the original dense model as reported in Tab.~\ref{tab:main_result}, whereas SALAD achieves comparable or superior results at a fraction of the resource cost. These findings underscore SALAD's tuning efficiency for sparse adaptation in large-scale video diffusion models.

\subsection{More Qualitative Results}
\label{app:qualitative_results}
\begin{figure*}[t!]
\centering
\includegraphics[width=1.0\columnwidth]{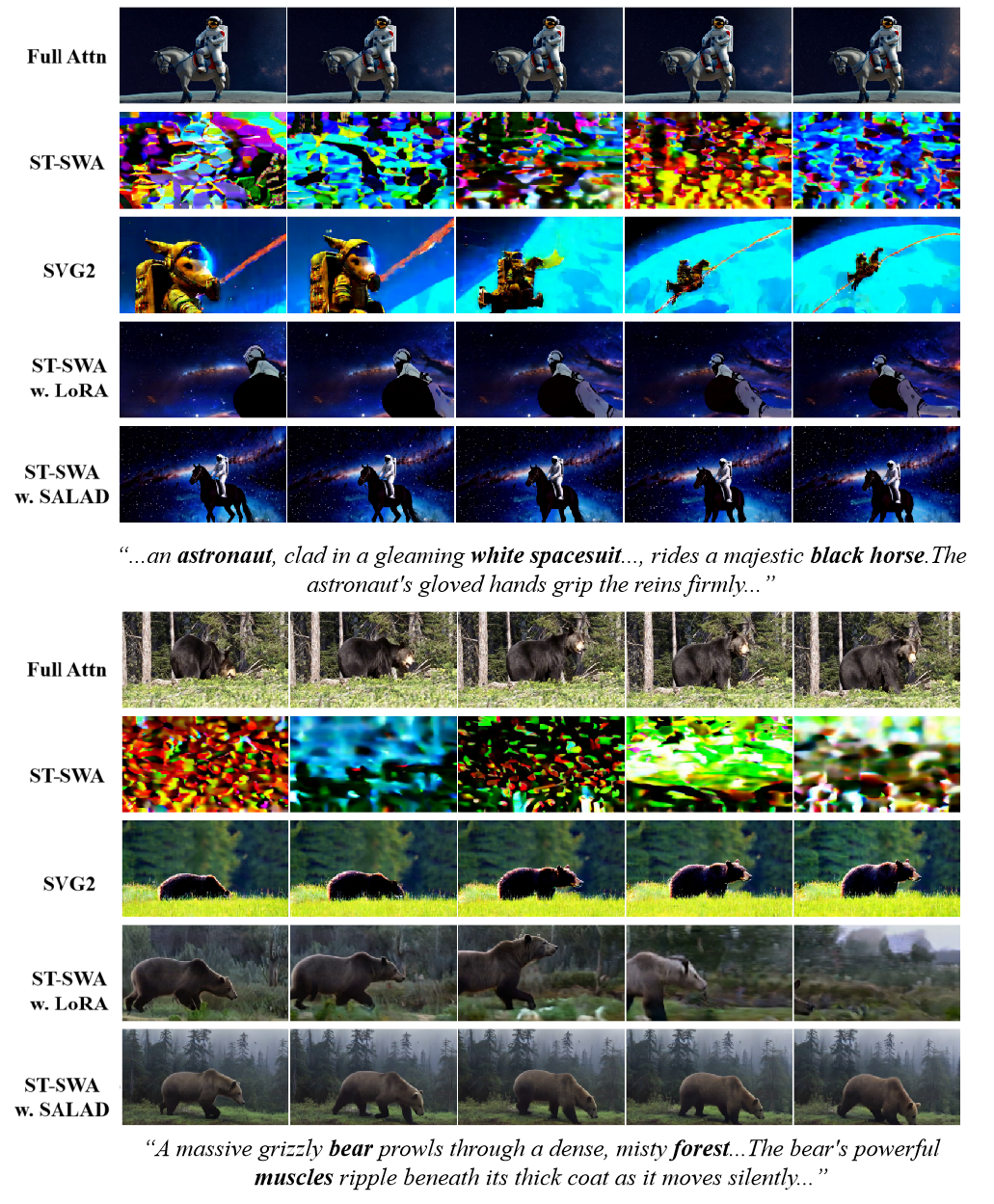}
\caption{\textbf{Generated Video Example of SALAD and other baselines.}}
\label{fig:qual_comp_1}
\end{figure*}

\begin{figure*}[t!]
\centering
\includegraphics[width=1.0\columnwidth]{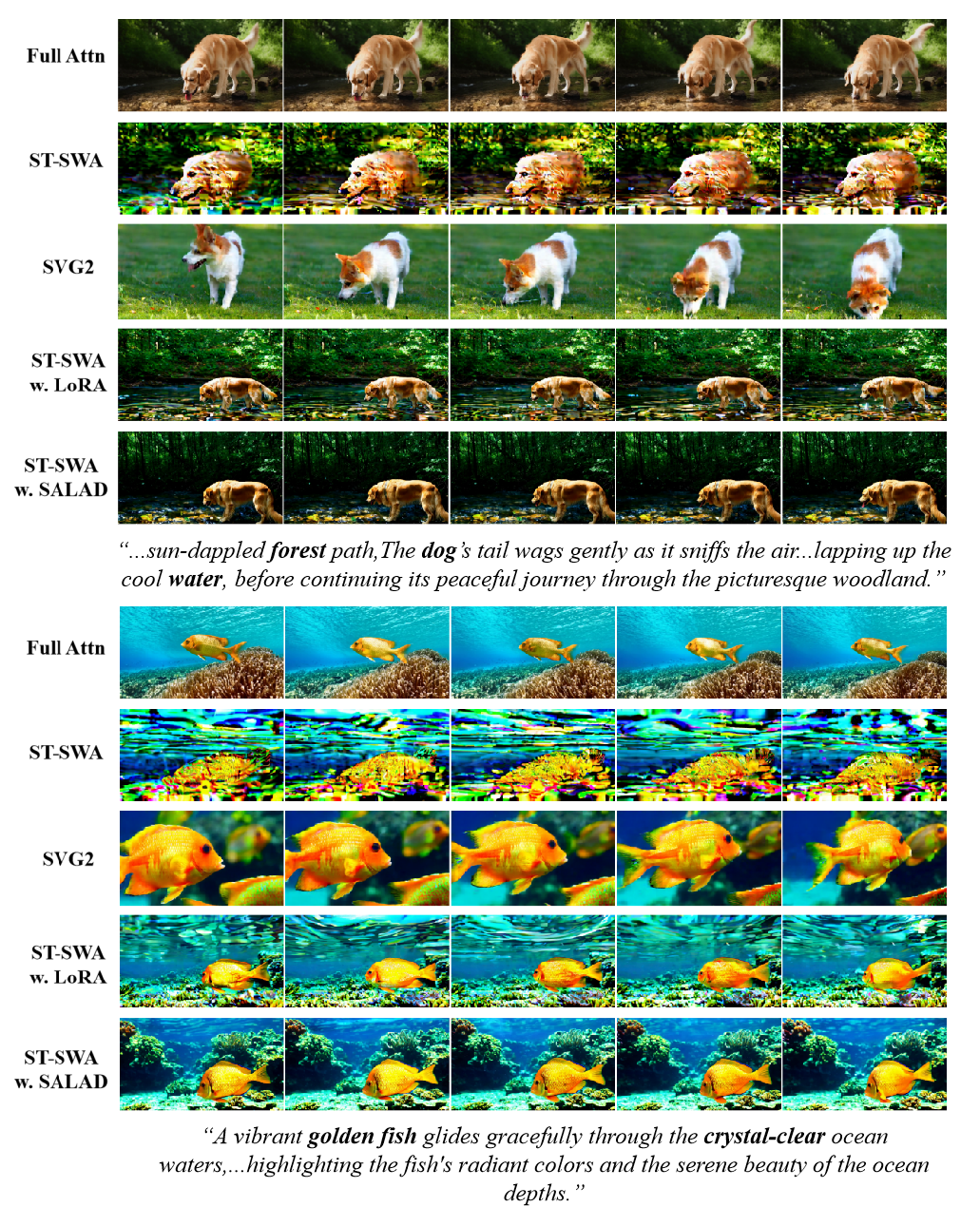}
\caption{\textbf{Generated Video Example of SALAD and other baselines.}}
\label{fig:qual_comp_2}
\end{figure*}

To further assess the visual quality and temporal consistency achieved by SALAD, we present additional qualitative comparisons against several representative baselines in this section. We evaluate SALAD with other baselines on Wan 2.1-1.3B. As illustrated in Figure~\ref{fig:qual_comp_1} and Figure~\ref{fig:qual_comp_2}, SALAD consistently preserves high-fidelity appearance details while generating temporally coherent motion, often matching or surpassing the visual quality produced by dense-attention counterparts. 

\end{document}